\documentclass[10pt,twocolumn,letterpaper]{article}
\pdfoutput=1
\usepackage[pagenumbers]{cvpr} 
\usepackage{makecell}
\usepackage{graphicx}
\usepackage{subcaption}

\usepackage[dvipsnames]{xcolor}
\usepackage{multirow}
\usepackage{booktabs} 
\usepackage{diagbox}
\usepackage{rotating}

\definecolor{cvprblue}{rgb}{0.21,0.49,0.74}
\usepackage[pagebackref,breaklinks,colorlinks,citecolor=cvprblue]{hyperref}

\def\paperID{2094} 
\def\confName{CVPR}
\def\confYear{2024}

\title{Partial Fine-Tuning: A Successor to Full Fine-Tuning for Vision Transformers}

\author{
Peng Ye\textsuperscript{\rm 1}$^\dagger$, Yongqi Huang\textsuperscript{\rm 1}$^\dagger$, Chongjun Tu\textsuperscript{\rm 1}, \\
Minglei Li\textsuperscript{\rm 1}, Tao Chen\textsuperscript{\rm 1}\thanks{Corresponding author.~~~$^\dagger$Equal Contribution}, Tong He\textsuperscript{\rm 2},  Wanli Ouyang\textsuperscript{\rm 2}\\
\textsuperscript{\rm 1}Fudan University, \textsuperscript{\rm 2}Shanghai AI Laboratory
}

\begin{document}
\maketitle
\begin{abstract}
Fine-tuning pre-trained foundation models has gained significant popularity in various research fields. Existing methods for fine-tuning can be roughly divided into two categories, namely Parameter-Efficient Fine-Tuning and High-Performance Fine-Tuning. The former aims at improving efficiency, while the latter focuses on enhancing performance. Beyond these methods, we demonstrate that Partial Fine-Tuning can be an innovative and promising direction capable of concurrently enhancing both efficiency and accuracy. We first validate eight manually-defined partial fine-tuning strategies across kinds of datasets and vision transformer architectures, and find that some partial fine-tuning strategies (e.g., ffn only or attention only) can achieve better performance with fewer tuned parameters than full fine-tuning, and selecting appropriate layers is critical to partial fine-tuning. Thus, we propose a novel fine-tuned angle metric to guide the selection of appropriate layers for partial fine-tuning, making it flexible to be adapted to various scenarios for more practicable partial fine-tuning. Additionally, we show that partial fine-tuning can serve as a new dimension for Model Soups, improving both the model performance and generalization with fewer tuned parameters. Comprehensive experiments on a wide range of datasets and models validate the great potential of partial fine-tuning.
\end{abstract}    
\section{Introduction}
\label{sec:intro}

\begin{figure}[t]
\centering
\includegraphics[width=0.90\linewidth]{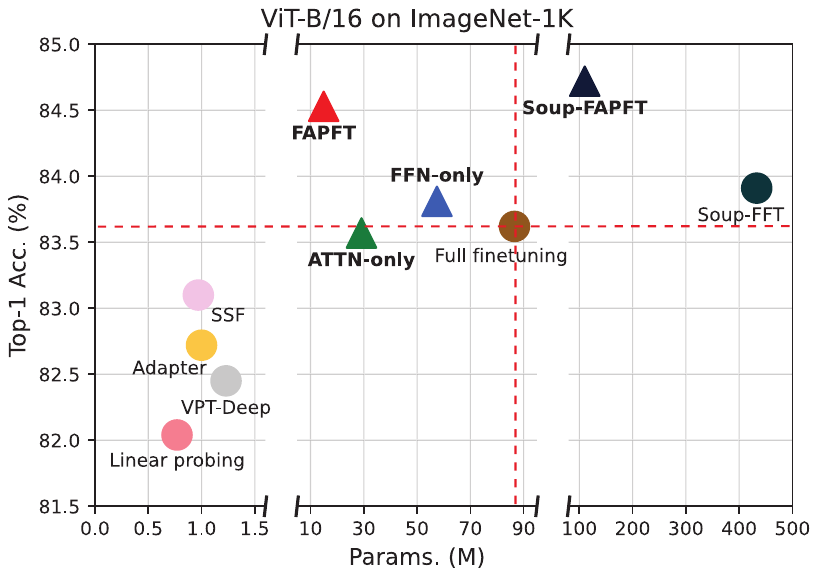}
\vspace{-6pt}
\caption{Performance comparisons of different kinds of fine-tuning methods, with a ViT-B/16 model pre-trained on ImageNet-21K and fine-tuned on ImageNet-1K. Our new perspective of partial fine-tuning (FAPFT, ATTN-only, and FFN-only) can improve both the performance and parameter efficiency of full fine-tuning, and be seamlessly combined with Model Soups (Soup-FAPFT).
More similar comparisons are shown in the Appendix.}
\label{fig:intro}
\vspace{-9pt}
\end{figure}

The integration of pre-training and fine-tuning has proven to be a successful approach in deep learning, facilitating the practical application of neural networks to a wide range of real-world problems. It has become a standard procedure in many domains and played a crucial role in advancing the state-of-the-art in various tasks~\cite{devlin2018bert,he2022masked,tang2023boosting,liang2023rethinking,ye2022stimulative,ye2023stimulative}. 
As the number of proposed pre-trained models continues to grow, how to effectively and efficiently adapt them to specific target tasks is attracting the increasing attention of researchers.

Existing fine-tuning methods can be roughly divided into two categories: Parameter-Efficient Fine-Tuning (PEFT) and High-Performance Fine-Tuning (HPFT). The former aims to improve the fine-tuning efficiency by freezing the pre-trained weights and fine-tuning only a few additional parameters~\cite{houlsby2019parameter,jia2022visual,hu2021lora}. 
The latter focuses on improving the model performance and generalization ability by combining model weights of multiple different fine-tuning~\cite{wortsman2022model,rame2022diverse,rame2023model}. 
Beyond these methods, a few recent works attempt to process partial parameters of the deep model.~\cite{shen2021partial} transfers partial knowledge of the pre-trained CNN model for better few-shot learning.~\cite{touvron2022three} finds that fine-tuning only the attention layers can adapt vision transformers. 
However, both methods focus on either specific tasks and networks or specific partial fine-tuning of specific architectures.
As a result, there remains a need for more comprehensive research and development of general methods on partial fine-tuning.



To comprehensively understand the role of partial fine-tuning, we first manually define eight different partial fine-tuning strategies,
and validate their effectiveness across kinds of datasets 
and vision transformer architectures.
The experimental results show that: 1) The performance of different partial fine-tuning strategies is affected by both the architectures and the datasets. 2) Partial fine-tuning of specific functional layers (e.g., ffn or attention) of vision transformers produces comparable or even better results than full fine-tuning. 3) The position of partially fine-tuned layers of vision transformers has a significant impact on the performance. For more details, please refer to Sec.~\ref{sec:experiments_findings}. These experiments and observations demonstrate the great potential of partial fine-tuning for achieving both high performance and parameter efficiency, while indicating that selecting appropriate layers is critical to improving the effectiveness and extending the utility of partial fine-tuning.

To this end, we develop a general and practicable partial fine-tining method based on a novel fine-tuned angle metric.
Firstly, we define the fine-tuned angle metric as the angle between pre-trained and fine-tuned weights of a specific layer since it can measure the training behavior during fine-tuning. Then, we compute and rank the fine-tuned angles of all layers under different kinds of fine-tuning configures (i.e., various hyper-parameters, epochs, and datasets), and find that different layers in the model may have different but constant effects when fine-tuning on a specific dataset. Further, we treat the layer with a large fine-tuned angle as the influential one and the layer with a small fine-tuned angle as the redundant one, and propose to fine-tune only the influential parts for challenging tasks to adapt most effectively, and fine-tune only the redundant parts for easy tasks to maintain the pre-trained knowledge maximally. As shown in Fig.~\ref{fig:intro}, although some manually-selected partial fine-tuning may already perform better than full fine-tuning, fine-tuned angle guided partial fine-tuning can surpass them on both the performance and parameter efficiency.

We conduct comprehensive experiments on kinds of datasets and models to validate the superiority of the proposed method. The results show that the proposed method can not only reduce the parameter number but also improve the performance.
Our novelties can be summarized as: 1) We in-depth investigate the role of partial fine-tuning for different vision transformers and datasets for the first time, and demonstrate its great potential for high-performance parameter-efficient fine-tuning. 2) We propose a novel fine-tuned angle guided partial fine-tuning approach, which is more effective and practicable by automatically selecting appropriate layers for fine-tuning in various scenarios. 3) We show that partial fine-tuning can serve as a new dimension for Model Soups, and is complementary to the common practice of using different training configures.

\section{Related Work}
\label{sec:formatting}

\subsection{Fine-tuning Methods}
The current research on model fine-tuning methods can be broadly divided into two main streams: Parameter-Efficient Fine-Tuning (PEFT) and High-Performance Fine-Tuning (HPFT). Beyond these two directions, we show that Partial Fine-Tuning can be a novel and promising direction. 

\textbf{Parameter-Efficient Fine-Tuning (PEFT)} is a recent technique that aims to adapt pre-trained models to new tasks while fine-tuning only a few additional parameters. 
In early attempts, a low-rank FFN module called Adapter is introduced by~\cite{houlsby2019parameter} and inserted between transformer layers, greatly improving the parameter efficiency of fine-tuning. 
In parallel, Prefix-tuning~\cite{li2021prefix} and VPT~\cite{jia2022visual} splice special prefix tokens into the input and only update the embedding of these unique prefix tokens when fine-tuning.
Besides, LoRA~\cite{hu2021lora} 
injects the rank-decomposition matrix into Transformer blocks as bypass modules and only finetunes these modules. 
Afterward, countless variations via these methods are proposed and applied to various 
tasks~\cite{lian2022scaling,hu2023llm,huang2023lorahub}.

\begin{figure*}
    \centering
    \includegraphics[width=0.92\linewidth]{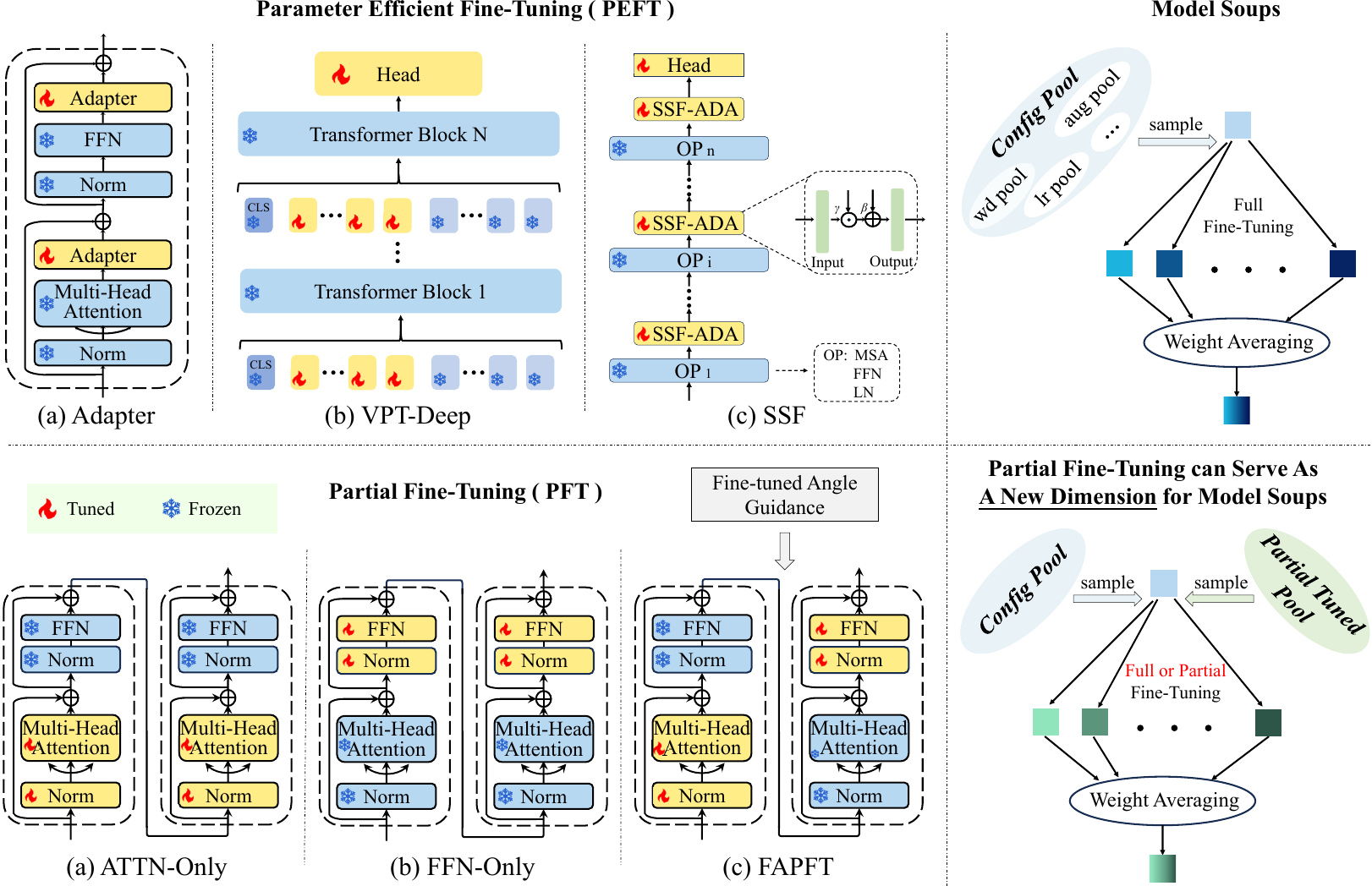}
    \vspace{-6pt}
    \caption{Illustrations of different fine-tuning methods. \textbf{Upper left}: Three representative Parameter Efficient Fine-Tuning methods, including Adapter, VPT-Deep, and SSF. \textbf{Bottom left}: The overall architecture of Partial Fine-Tuning methods, with (a) and (b) representing all attentions and all ffns fine-tuning, and (c) representing fine-tuned angle guided partial fine-tuning. \textbf{Upper right}: A brief illustration of the Model Soups method. \textbf{Bottom right}: An illustration of how Partial Fine-Tuning serves as a new dimension for Model Soups.}
    \label{fig:main_comparisons}
    \vspace{-2mm}
\end{figure*}

\textbf{High-Performance Fine-Tuning (HPFT)} focuses on maximizing performance by utilizing model weights of multiple fine-tuning. Model Soups~\cite{wortsman2022model} averages weights obtained by fine-tuning the same pre-trained model with different configures, improving performance without additional inference costs. DiWA~\cite{rame2022diverse} employs diverse weight averaging for better generalization, especially for out-of-distribution (OOD) data. Model Ratatouille~\cite{rame2023model} reuses the weights obtained by first fine-tuning on multiple auxiliary tasks and then fine-tuning on the target task, further enhancing the OOD generalization ability.

Besides, recent studies processing partial parameters of deep models have brought new insights into fine-tuning. For better few-shot learning,~\cite{shen2021partial} transfers partial knowledge from the base set to the novel set by setting different learning rates for different layers of the pre-trained CNN model.~\cite{touvron2022three} indicates that merely fine-tuning the weights of attention layers can already adapt ViTs to different resolutions and classification tasks.
However, the former places its emphasis on the few-shot task and CNN model, and the latter focuses on the specific partial fine-tuning of ViTs.
Differently, we present the first comprehensive study of partial fine-tuning, validating the effectiveness of various partial fine-tuning on a range of datasets and models. We also introduce a novel fine-tuned angle guided partial fine-tuning strategy and find that partial fine-tuning can serve as a new dimension for Model Soups. We show the details in Fig.~\ref{fig:main_comparisons}. 

\subsection{Angle Metric and Its Applications}
Recently, the deep learning community has recognized the significance of angle metric in evaluating the training behavior of neural networks. Several studies~\cite{arora2018theoretical,li2019exponential} have theoretically demonstrated that the angle of model weights provides an accurate measure of weight updates.
In~\cite{carbonnelle2019layer}, the angle metric is proposed, which is defined as the angle between the initialized weights and that of a well-trained network and is used to assess the network's generalization capability. Further, ABS~\cite{hu2020angle} introduces the angle metric to represent the whole model performance and utilizes it to shrink the search space of Neural Architecture Search (NAS), RLNAS~\cite{zhang2021neural} employs such angle metric to search for the optimal architecture, and AngleLoss~\cite{yang2023revisiting} discusses the impact on NAS when applying the angle metric to the feature extraction and prediction layers respectively. Unlike the above methods that use the angle metric to indicate the performance or generalization of different architectures or operators, we explore how to leverage it to guide partial fine-tuning. Besides, all the above methods measure the angle between the initialized and well-trained model weights, while we explore the angle of each layer between pre-trained and fine-tuned model weights. Since such a finetuned angle metric has not been studied before, we conduct some empirical studies and show some interesting findings.

\begin{figure*}[htbp]
  \centering
  \begin{subfigure}[b]{0.32\textwidth}
    \includegraphics[width=\textwidth]{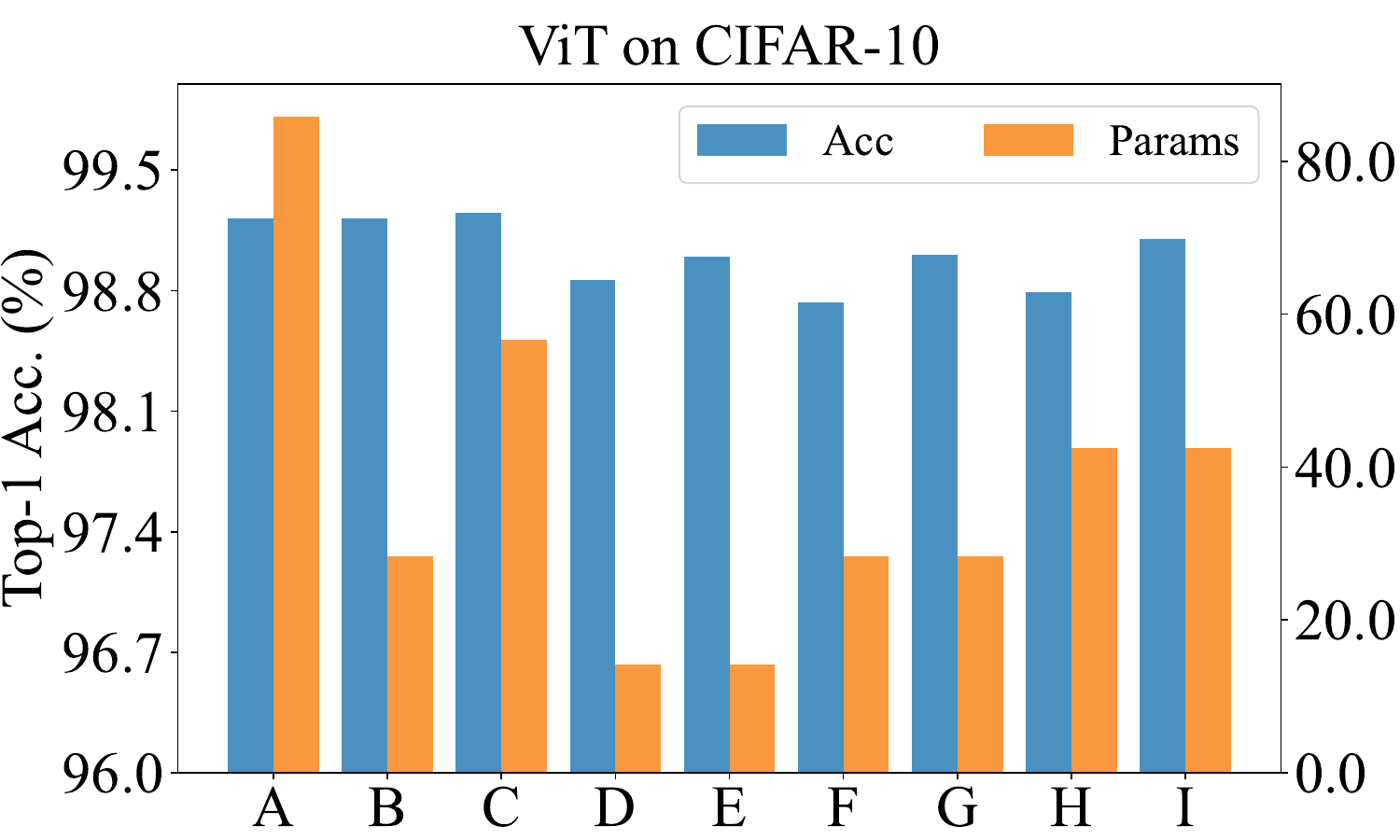}
    \label{subfig:1}
  \end{subfigure}
  \begin{subfigure}[b]{0.32\textwidth}
    \includegraphics[width=\textwidth]{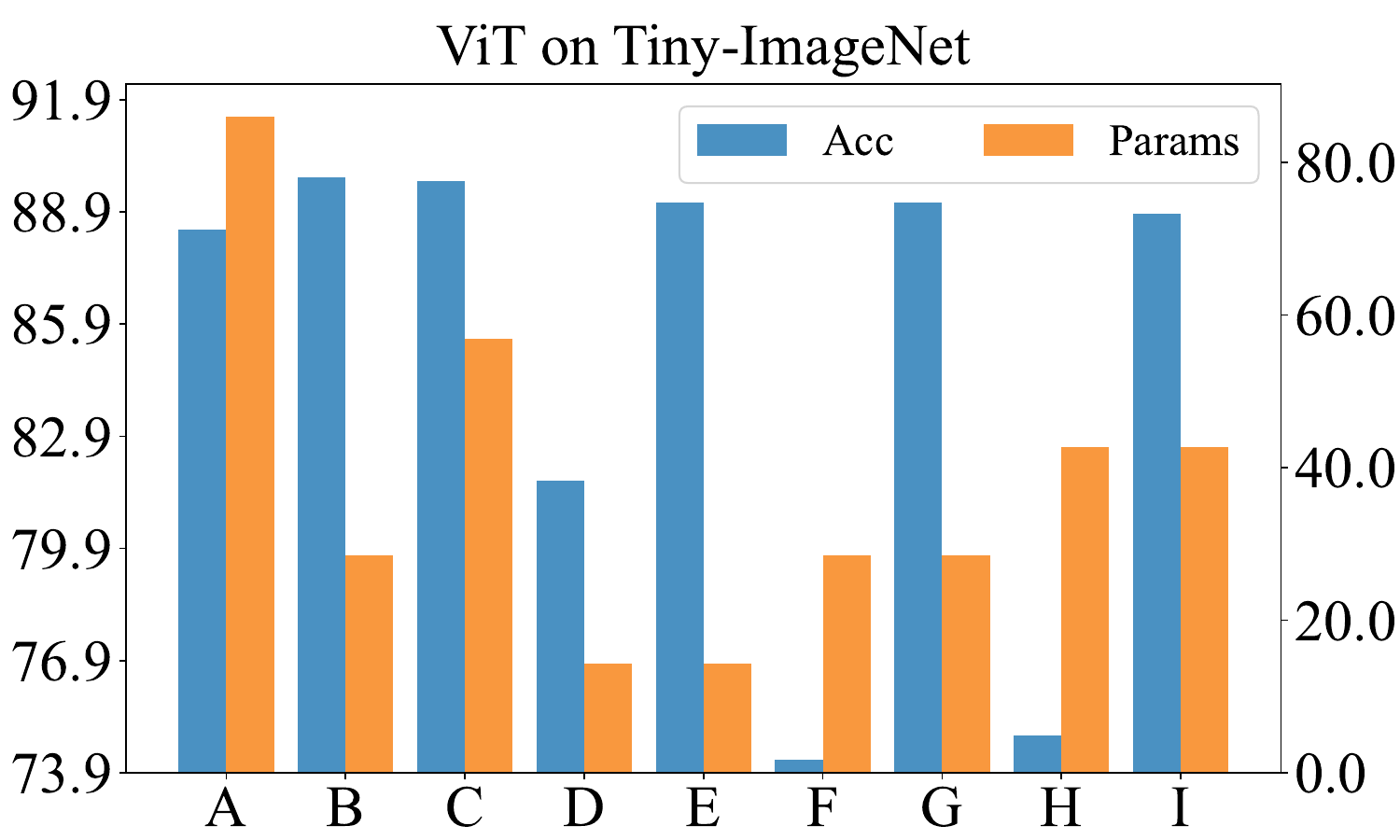}
    \label{subfig:2}
  \end{subfigure}
  \begin{subfigure}[b]{0.32\textwidth}
    \includegraphics[width=\textwidth]{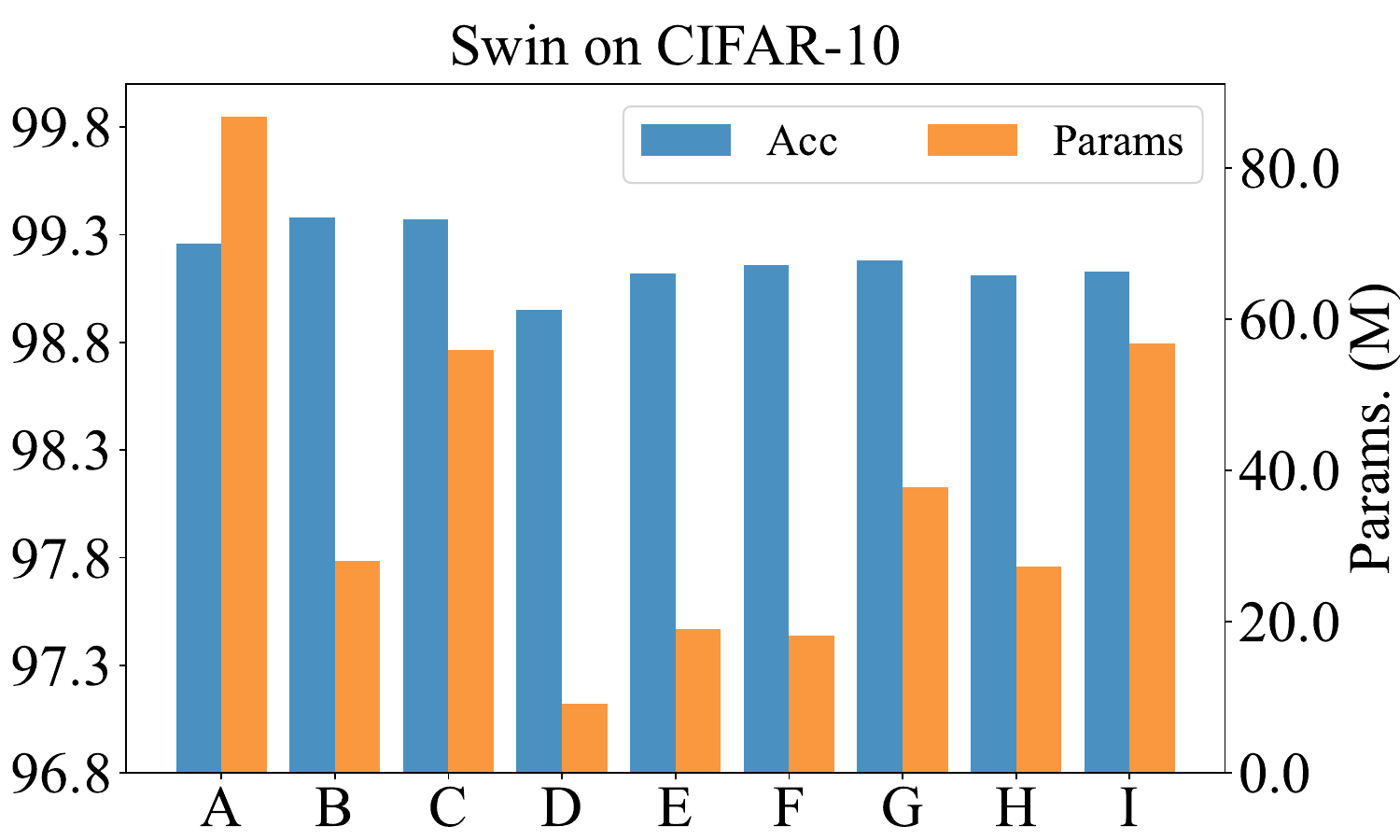}
    \label{subfig:3}
  \end{subfigure}\vspace{-4mm}
  
  \begin{subfigure}[b]{0.32\textwidth}
    \includegraphics[width=\textwidth]{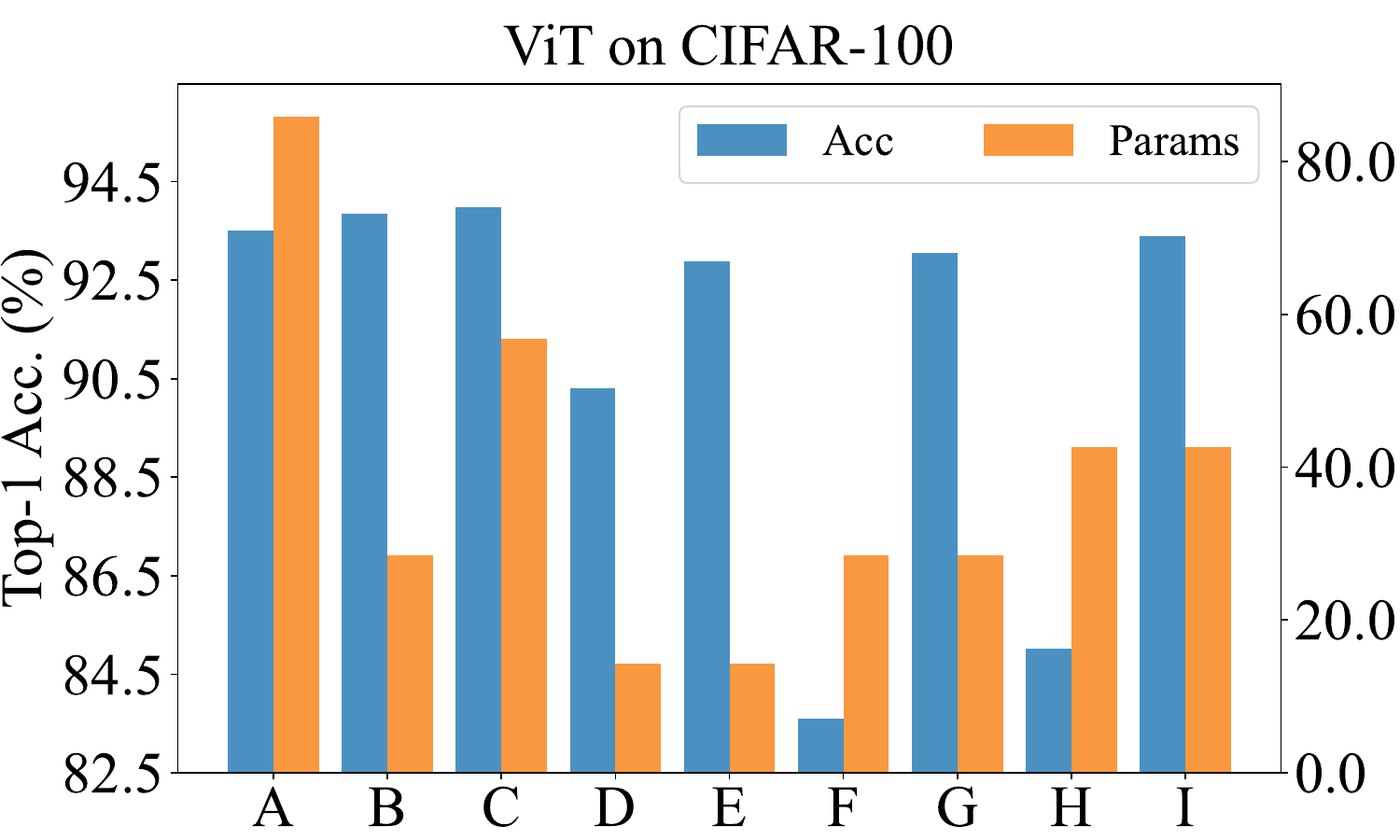}
    \label{subfig:4}
  \end{subfigure}
  \begin{subfigure}[b]{0.32\textwidth}
    \includegraphics[width=\textwidth]{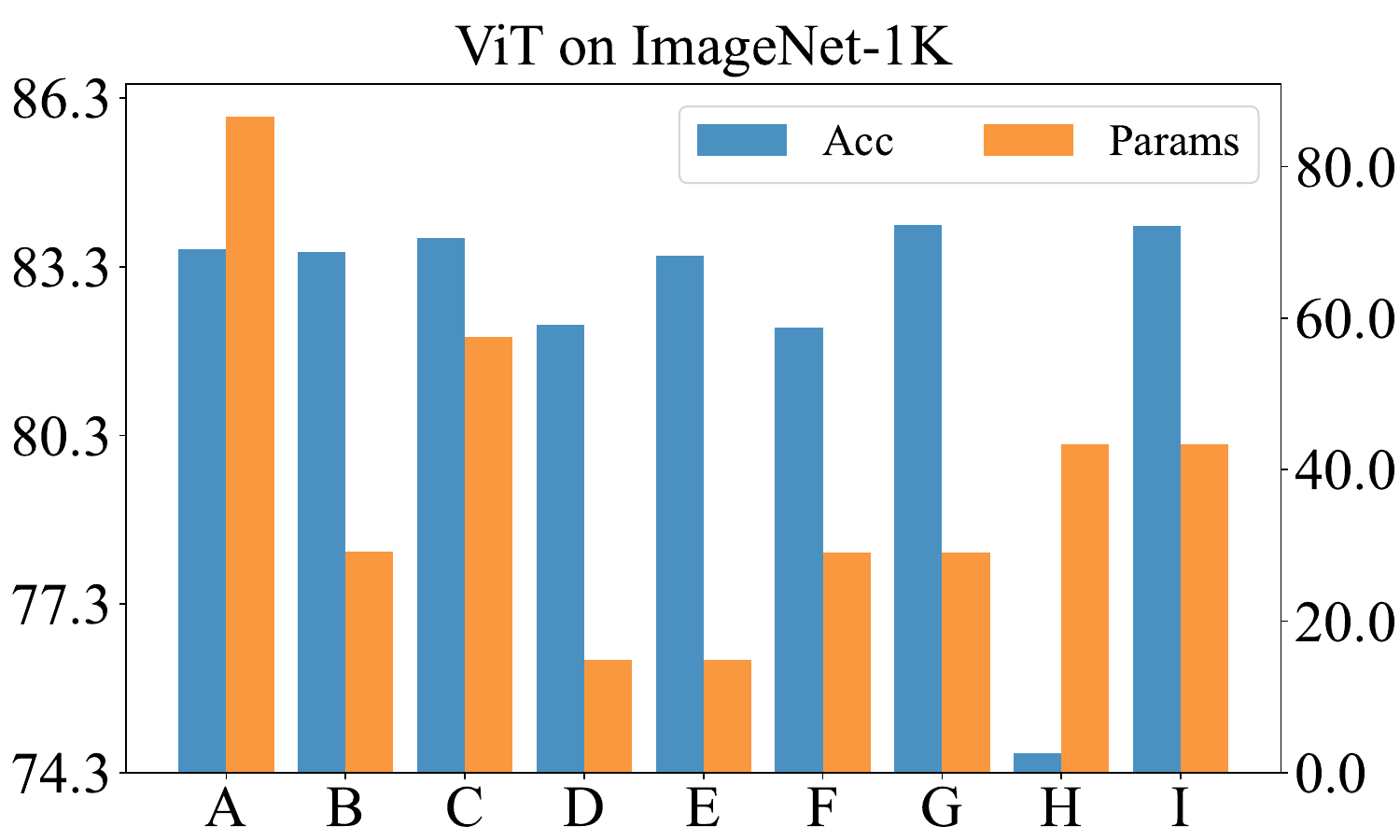}
    \label{subfig:5}
  \end{subfigure}
  \begin{subfigure}[b]{0.32\textwidth}
    \includegraphics[width=\textwidth]{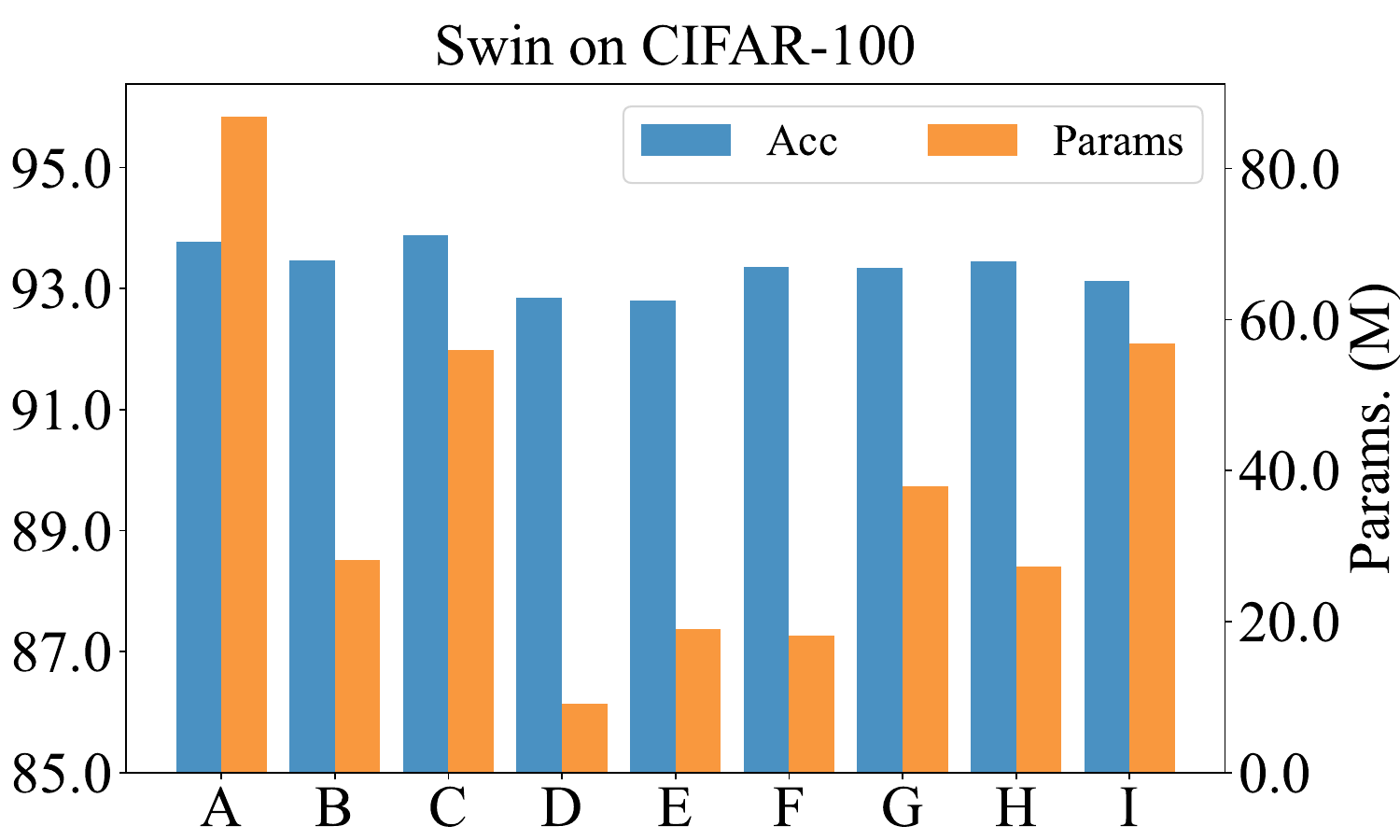}
    \label{subfig:6}
  \end{subfigure}\vspace{-6mm}
  \caption{Accuracy and parameter comparisons of eight manually-designed partial fine-tuning and full fine-tuning on various models and datasets.
  A on the horizontal axis of all sub-figures represents full fine-tuning, while B, C, D, E, F, G, H, I represent only fine-tuning all attentions, all ffns, first half attentions, last half attentions, first half ffns, last half ffns, first half blocks, and last half blocks, respectively.}
  \label{fig:exploration}
  \vspace{-2mm}
\end{figure*}

\section{Method}

\subsection{Partial Fine-Tuning Is All You Need} \label{sec:experiments_findings}

To gain 
a more comprehensive understanding of partial fine-tuning, we conduct several exploratory experiments across various datasets (CIFAR-10, CIFAR-100, Tiny-ImageNet, and ImageNet-1K) and vision transformer architectures (ViT, and Swin Transformer). In each task setting, we compare the performance and number of trainable parameters between full fine-tuning and eight different partial fine-tuning strategies. The results are shown in ~\cref{fig:exploration}, from which we can conclude three insightful observations. 

\textbf{Insightful Observation 1:} The performance of different partial fine-tuning strategies is influenced by both the architecture and dataset. First, the optimal partial fine-tuning strategy differs across different architectures and datasets. Second, different architectures and datasets have different sensitivities to partial fine-tuning. For example, the performance differences for all partial fine-tuning strategies are small for ViT on CIFAR-10, Swin on CIFAR-10, and Swin on CIFAR-100, while relatively large for other strategies.

\textbf{Insightful Observation 2:} 
Partial fine-tuning of specific functional layers of vision transformers yields comparable or even better results than full fine-tuning. 
This is particularly evident when examining the performance of fine-tuning only the attention layers or the ffn layers,
which substantially reduce the trainable parameters while preserving satisfactory results. This phenomenon also validates the great potential of partial fine-tuning.

\textbf{Insightful Observation 3:} The position of the fine-tuned layers of various vision transformers plays a significant role in the performance of partial fine-tuning. For example, with the same trainable parameters, fine-tuning only the latter half of layers consistently performs better than fine-tuning only the first half of layers, which is especially apparent when considering the fine-tuned results of ViT on CIFAR-100, Tiny-ImageNet, and ImageNet-1K datasets.

In light of these observations, a strategy to select appropriate layers for fine-tuning in various scenarios becomes crucial to extending the utility of partial fine-tuning. 

\begin{figure*}
    \centering
    \includegraphics[width=0.92\linewidth]{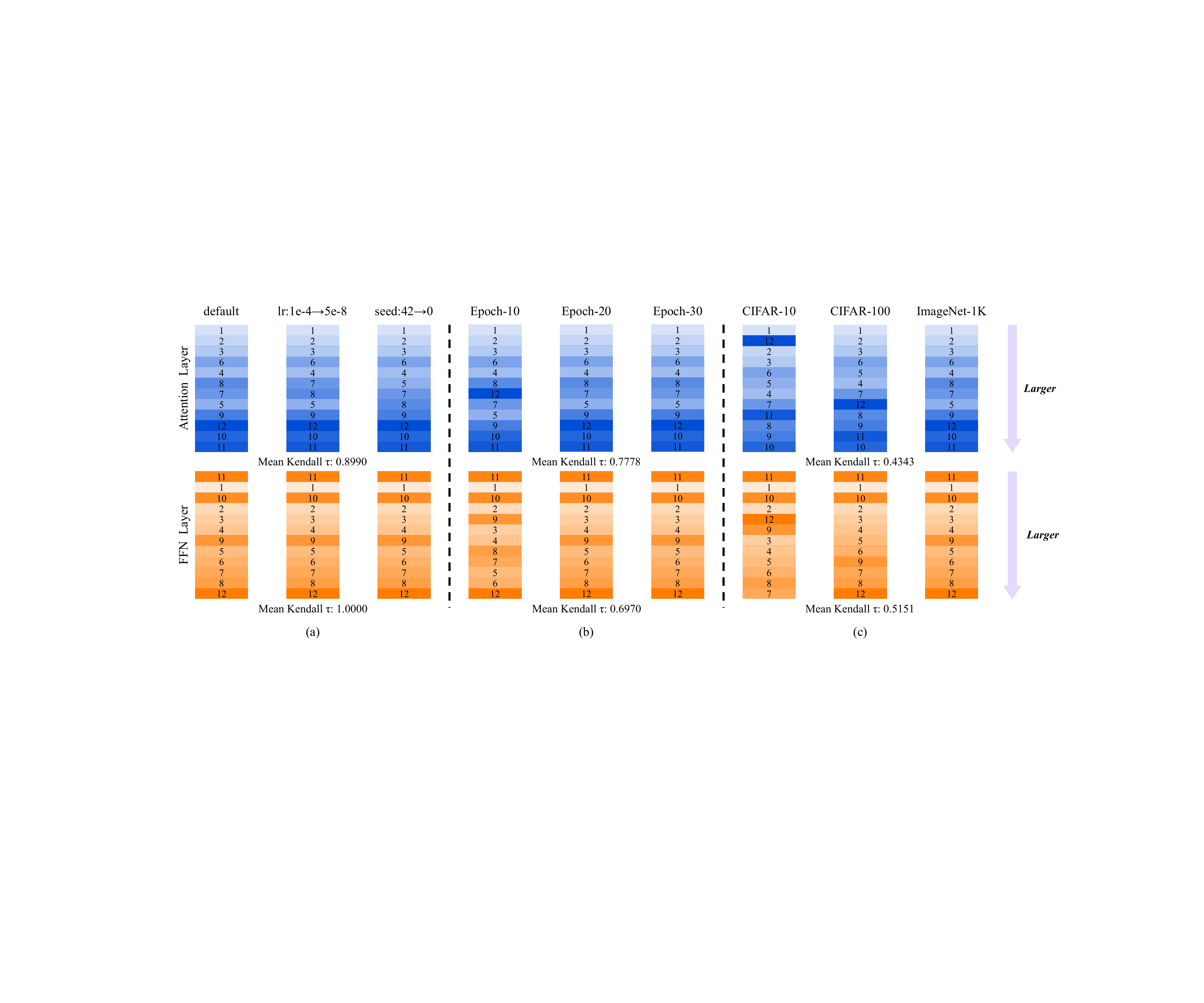}
    \vspace{-3mm}
    \caption{Visualization of the ranking of fine-tuned angles for all attention or FFN layers on ViT-B/16. We explore the influence of various training (a) hyper-parameters and (b) epochs on ImageNet-1K, and (c) datasets. Mean Kendall correlations for angle rankings are shown. For each column, moving from top to bottom corresponds to an increase in layer depth. Numbers and colors represent angle rankings: a smaller number and a lighter color indicate a higher rank (larger value) of the fine-tuned angle. 
    }
    \label{fig:angle_rank}
    \vspace{-3mm}
\end{figure*}

\subsection{Fine-tuned Angle guided Partial Fine-Tuning}
Based on the observations in Sec.~\ref{sec:experiments_findings}, when proper parts are chosen for kinds of datasets and architectures, partial fine-tuning can achieve comparable performance with fewer parameters than full fine-tuning. Thus, we shift our attention to how to select proper parts for partial fine-tuning.

Since the angle metric is widely used for evaluating the training behavior of neural networks, we introduce it to measure the impact of various fine-tuning configures on different layers of the given model. Original angle metric, which is utilized for either indicating the generalization capacity of a model~\cite{arora2018theoretical,li2019exponential} or ranking candidate architectures in NAS~\cite{hu2020angle,zhang2021neural,yang2023revisiting}, converts the weights of a whole network into a one-dimensional vector, and computes the angle between the weight vectors before and after training. In detail, given a model $M$, let $\boldsymbol{W}_{\mathbf{0}}$ and $\boldsymbol{W}_{\mathbf{t}}$ denotes its initialization weight vector and trained weight vector respectively, then the original angle metric can be computed as
\begin{equation}
\theta(\boldsymbol{M})=\arccos \left(\frac{\boldsymbol{W}_{\mathbf{0}} \cdot \boldsymbol{W}_{\boldsymbol{t}}}{\|\boldsymbol{W}_{\mathbf{0}}\|_{2}\|\boldsymbol{W}_{\boldsymbol{t}}\|_{2}}\right)
\end{equation}

\noindent
\textbf{Fine-tuned Angle Metric.} As we want to investigate the training behavior of each layer when fine-tuning the model, which has not been explored before, we first define a new fine-tuned angle metric. For a specific layer $L$ in the model, we convert its pre-trained weights and fine-tuned weights into a one-dimensional vector respectively, denoted as $\boldsymbol{W}_{\mathbf{p}}^L$ and $\boldsymbol{W}_{\mathbf{f}}^L$, then the fine-tuned angle metric are calculated as
\begin{equation}
\theta(\boldsymbol{L})=\arccos \left(\frac{\boldsymbol{W}_{\mathbf{p}}^L \cdot \boldsymbol{W}_{\boldsymbol{f}}^L}{\|\boldsymbol{W}_{\mathbf{p}}^L\|_{2}\|\boldsymbol{W}_{\boldsymbol{f}}^L\|_{2}}\right)
\label{eq:finetuned_angle}
\end{equation}

\noindent
\textbf{Fine-tuned Angle of Various Configures.} We then compute the fine-tuned angles of different layers of the model under different fine-tuning configures.
For a comprehensive study, we explore the effect of kinds of fine-tuning hyperparameters, iterations, and datasets. The results are shown in Fig.~\ref{fig:angle_rank}. As we can see, the ranking of the fine-tuning angles of all layers is surprisingly consistent under different fine-tuning hyperparameters and iterations (as shown in Fig.~\ref{fig:angle_rank} (a) and (b)), while different fine-tuning datasets result in very different rankings of the fine-tuning angles of all layers (as shown in Fig.~\ref{fig:angle_rank} (c)). Such results reveal that different layers in the model may have different but constant effects on the fine-tuning process for a specific dataset, which further inspires us to consider using such fine-tuned angle as a kind of guidance for partial fine-tuning.

\noindent
\textbf{Fine-tuned Angle guided Partial Fine-Tuning.} Recent works have investigated the difference value between the pre-trained model weights and the fine-tuned one and further regarded the large difference value as the influential one and the small difference value as the redundant one~\cite{ilharco2022editing,yadav2023resolving}. Inspired by this, we also treat the layer with a large fine-tuned angle as the influential layer and the layer with a small fine-tuned angle as the redundant layer. Further, for challenging tasks, we can fine-tune only the influential parts (layers with large fine-tuned angles) to adapt most effectively. For easy tasks, we can fine-tune only the redundant parts (layers with small fine-tuned angles) to maintain the pre-trained knowledge maximally. In the experiments section, we show the effectiveness of such a strategy, which performs much better than manually selected partial fine-tuning strategies on kinds of datasets and architectures.

\begin{table*}[t]
	\centering
	\setlength{\tabcolsep}{2.8pt} 
	\scalebox{0.90}{\begin{tabular}{c|cc|cc|cc|cc|cc|cc|cc}
			\toprule
			Model &  \multicolumn{4}{c|}{ViT-B/16~\cite{dosovitskiy2020image}} & \multicolumn{4}{c|}{Swin-B~\cite{liu2021swin}} & \multicolumn{4}{c|}{ConvNeXt-B~\cite{liu2022convnet}} & \multicolumn{2}{c}{AS-MLP-B \cite{lian2021mlp}} \\ \midrule 
			\diagbox{Method}{Dataset} & \rotatebox{90}{CIFAR-100~} & \rotatebox{90}{Params. (M)~} &  \rotatebox{90}{ImageNet-1K~}  & \rotatebox{90}{Params. (M)~} & \rotatebox{90}{CIFAR-100~} &  \rotatebox{90}{Params. (M)~} &  \rotatebox{90}{ImageNet-1K~}  & \rotatebox{90}{Params. (M)~} &  \rotatebox{90}{CIFAR-100~} &  \rotatebox{90}{Params. (M)~} & \rotatebox{90}{ImageNet-1K~}  & \rotatebox{90}{Params. (M)~} & \rotatebox{90}{CIFAR-100~}  & \rotatebox{90}{Params. (M)~}  \\ \midrule
			Full fine-tuning  & 93.51 & 85.88 & 83.62 & 86.57 & 93.77 & 86.85 & \underline{85.07} & 87.77 & \underline{94.04} & 87.67 & \textbf{85.49} & 88.59 & \underline{90.04} & 86.83 \\ 
			Linear probing   & 88.70 & 0.08 &  82.04 & 0.77   & 89.27 &  0.10 & 83.25 &  1.03 & 89.20 &  0.10 &  84.05 & 1.03 & 79.04 & 0.10 \\ \midrule 
			Adapter \cite{houlsby2019parameter} & 93.34 & 0.31 & 82.72 & 1.00  & 92.49  & 0.33 & 83.82 & 1.26 & 92.86  & 0.45 & 84.49 & 1.37 & 88.01 &0.33 \\
			VPT-Deep \cite{jia2022visual} & 93.17 & 0.54 & 82.45   & 1.23 & 92.62  & 0.70  & 83.44  & 1.63 &  - & - &  - &  - & - & - \\ 
			SSF \cite{lian2022scaling} & \underline{93.99} & 0.28 & 83.10 &  0.97 & 93.06 & 0.37   & 84.40 & 1.29 & 93.45 & 0.36 & 84.85  & 1.28& 88.28 & 0.37\\ \midrule
            ATTN-Only & 93.84 & 28.44 & 83.57 & 29.14 & 93.46 & 28.16 & 84.58 & 29.08 & - & - & - & - & - & - \\ 
            FFN-Only & 93.98 & 56.76 & \underline{83.81} & 57.46 & \underline{93.88} & 56.02 & 84.88 & 56.95 & - & - & - & - & - & - \\
            FAPFT \textbf{(ours)} & \textbf{94.30} & 49.69 & \textbf{84.53} & 14.95 & \textbf{94.07} & 33.61 & \textbf{85.17} & 42.01 & \textbf{94.05} & 45.19 & \underline{85.38} & 39.76 & \textbf{90.74} & 46.21 \\ 
			\bottomrule
		\end{tabular}
	}
    \vspace{-2mm}
	\caption{Performance comparisons of diverse fine-tuning approaches across different model architectures on CIFAR-100 and ImageNet-1K. Except for the AS-MLP-B model, which is pre-trained on ImageNet-1K, the other models, including ViT-B/16, Swin-B, and ConvNeXt-B, are pre-trained on ImageNet-21K. Each partial fine-tuning utilizes the same hyper-parameters as full fine-tuning, while others do not.}
	\label{table:imagenet}
\end{table*}

\begin{table*}[t]
	\centering
	\scalebox{0.9}{\begin{tabular}{c|c|c|c|c|c|c|c}
        \toprule
        \diagbox{Method}{Dataset}  & \makecell[c]{~CUB-200~ \\ -2011} & ~NABirds~ & \makecell[c]{~Oxford~ \\ Flowers}  & \makecell[c]{~Stanford~ \\ Dogs}  & \makecell[c]{~Stanford~ \\ Cars}  & ~~Mean~~ & \makecell[c]{~Params. \\ (M)}   \\ \midrule
        Full fine-tuning & 87.30 & 82.70 & 98.80 & 89.40 & 84.50 & 88.54 & 85.98   \\ 
        Linear probing & 85.30 & 75.90 & 97.90 & 86.20 & 51.30 & 79.32 &  0.18 \\ \midrule
        Adapter \cite{houlsby2019parameter} & 87.10 & \underline{84.30} & 98.50 & 89.80 & 68.60 & 85.67 & 0.41 \\ 
        VPT-Deep \cite{jia2022visual} & \underline{88.50} & 84.20 & \underline{99.00} & \underline{90.20} & 83.60 & 89.11 & 0.85 \\
        SSF \cite{lian2022scaling} & 82.70 & \textbf{85.90} & 98.50 & 87.70 & 82.60 & 87.48 & 0.39 \\ \midrule
        ATTN-Only & 87.95 & 83.52 & 98.93 & 89.52 & \underline{87.48} & \underline{89.48} & 29.14 \\
        FFN-Only & 86.23 & 83.24 & 98.81 & 90.14 & 83.75 & 88.43 & 57.46 \\
        FAPFT \textbf{(ours)} & \textbf{88.68} & 83.79 & \textbf{99.04} & \textbf{91.20} & \textbf{88.15} & \textbf{90.17} & 30.69 \\ \bottomrule
    \end{tabular}
  }
    \vspace{-2mm}
	\caption{Performance comparisons of diverse fine-tuning strategies across five FGVC datasets, using ViT-B/16 model pre-trained on ImageNet-21K as the backbone. To guarantee a fair comparison, the results for the SSF method presented in the table are replicated from \cite{dong2023efficient}, which uses the same fundamental data augmentations as others. Note that PEFT methods (Adapter, VPT-Deep, and SSF) employ a grid search for each task to optimize hyper-parameters,
 while partial fine-tuning methods 
 use identical hyper-parameters as 
 full fine-tuning.}
	\label{table:fgvc}
\end{table*}

\begin{table*}[h]
	\centering
        \subfloat[ViT-B/16 on ImageNet-1K]{
		\setlength{\tabcolsep}{2pt}
		\scalebox{0.7}{
			\begin{tabular}{c|cc|cc}
				\toprule
                Method & \multicolumn{2}{c|}{FFT} & \multicolumn{2}{c}{FAPFT} \\ \midrule
				Exp. &  Acc.   & Params. &  Acc.   & Params. \\ \midrule
				run1 & 83.61 & 86.57 & 84.25 & 7.86\\
				run2 & 83.66 & 86.57 & 84.53 & 14.95\\
				run3 & 83.62 & 86.57 & 84.42 & 22.03\\
				run4 & 83.45 & 86.57 & 84.37 & 29.12\\
				run5 & 83.86 & 86.57 & 84.20 & 36.21\\ \midrule
                Soup & 83.91 & 432.9 & 84.72 & 110.2 \\
				\bottomrule
		\end{tabular}}
	}
	\subfloat[ViT-B/16 on CIFAR-100]{
		\setlength{\tabcolsep}{2pt}
		\scalebox{0.7}{
			\begin{tabular}{c|cc|cc}
				\toprule
                Method & \multicolumn{2}{c|}{FFT} & \multicolumn{2}{c}{FAPFT} \\ \midrule
				Exp. &  Acc.   & Params. &  Acc.   & Params. \\ \midrule
				run1 & 93.95 & 85.88 & 93.53 & 14.25\\
				run2 & 93.45 & 85.88 & 93.69 & 21.34\\
				run3 & 93.81 & 85.88 & 94.09 & 28.43\\
				run4 & 93.51 & 85.88 & 94.15 & 35.52\\
				run5 & 93.14 & 85.88 & 94.30 & 49.69\\ \midrule
                Soup & 94.05 & 429.4 & 94.37 & 146.2 \\
				\bottomrule
		\end{tabular}}
	}
	\subfloat[Swin-B on ImageNet-1K]{
		\setlength{\tabcolsep}{2pt}
		\scalebox{0.7}{
			\begin{tabular}{c|cc|cc}
				\toprule
                Method & \multicolumn{2}{c|}{FFT} & \multicolumn{2}{c}{FAPFT} \\ \midrule
				Exp. &  Acc.   & Params. &  Acc.   & Params. \\ \midrule
				run1 & 84.98 & 87.77 & 85.04 & 29.39\\
				run2 & 84.99 & 87.77 & 85.12 & 32.54\\
				run3 & 84.93 & 87.77 & 85.17 & 42.01\\
				run4 & 85.01 & 87.77 & 85.17 & 51.47\\
				run5 & 85.07 & 87.77 & 85.11 & 57.78\\ \midrule
                Soup & 85.18 & 438.9 & 85.25 & 213.2 \\
				\bottomrule
		\end{tabular}}
	}
	\subfloat[Swin-B on CIFAR-100]{
		\setlength{\tabcolsep}{2pt}
		\scalebox{0.7}{
			\begin{tabular}{c|cc|cc}
				\toprule
                Method & \multicolumn{2}{c|}{FFT} & \multicolumn{2}{c}{FAPFT} \\ \midrule
				Exp. &  Acc.   & Params. &  Acc.   & Params. \\ \midrule
				run1 & 93.68 & 86.85 & 94.00 & 32.62 \\
				run2 & 93.57 & 86.85 & 94.07 & 33.61 \\
				run3 & 93.65 & 86.85 & 94.03 & 35.78 \\
				run4 & 93.77 & 86.85 & 94.02 & 36.77 \\
				run5 & 93.56 & 86.85 & 93.86 & 45.23 \\ \midrule
                Soup & 93.91 & 434.3 & 94.14 & 184.0 \\
				\bottomrule
		\end{tabular}}
	}
        \vspace{-2mm}
	\caption{Comparisons of full fine-tuning (FFT) based and FAPFT based Model Soups. We present the results of 5 individual runs for both methods, subsequently amalgamating them into Model Soups. The final row provides the final performance and the total parameters. Our FAPFT-based soup shows better performance and fewer parameters than FFT-based soup across various models and datasets.
 }
	\label{table:model-soup}
\end{table*}

\begin{table*}[h]
	\centering
	\subfloat[ViT-B/16 on Robustness and OOD Datasets]{
		\setlength{\tabcolsep}{7pt}
	  \scalebox{0.7}{\begin{tabular}{c|c|c|c|c|c}
			\toprule
			\diagbox{Method}{Dataset}& IN-1K ($\uparrow$) & IN-A ($\uparrow$) & IN-R ($\uparrow$) & IN-C ($\downarrow$) & Params. \tabularnewline \midrule
			Full fine-tuning & 83.62 & 37.36 & 53.75 & 43.40 & 86.57 \tabularnewline \midrule
            ATTN-Only & 83.57 & 42.33 & 55.51 & 42.16 & 29.14 \tabularnewline
            FFN-Only & 83.81 & 40.35 & 54.47 & 42.77 & 56.76 \tabularnewline
            FAPFT \textbf{(ours)}  & 84.53 & 45.00 & 55.04 & 41.38 & 14.95 \tabularnewline  \midrule
            Soup-FFT & 83.91 & 41.40 & 55.39 & 41.84 & 432.9\tabularnewline
            Soup-FAPFT & 84.72 & 46.67 & 56.23 & 40.08 & 110.2\tabularnewline
            \bottomrule
		\end{tabular}
	}
	}
	\subfloat[Swin-B on Robustness and OOD Datasets]{
		\setlength{\tabcolsep}{7pt}
	  \scalebox{0.7}{\begin{tabular}{c|c|c|c|c|c}
			\toprule
			\diagbox{Method}{Dataset}& IN-1K ($\uparrow$) & IN-A ($\uparrow$) & IN-R ($\uparrow$) & IN-C ($\downarrow$) & Params. \tabularnewline \midrule
		  Full fine-tuning & 85.07 & 48.39 & 58.39 & 44.23 & 87.77  \tabularnewline \midrule
            ATTN-Only & 84.58 & 50.16 & 57.82 & 44.36 & 29.08 \tabularnewline
            FFN-Only & 84.88 & 49.81 & 58.63 & 43.93 & 56.95 \tabularnewline
            FAPFT \textbf{(ours)}  & 85.17 & 50.17 & 57.20 & 43.91 & 42.01 \tabularnewline  \midrule
            Soup-FFT & 85.18 & 50.99 & 58.98 & 43.06 & 438.9\tabularnewline
            Soup-FAPFT & 85.25 & 51.71 & 58.22 & 42.80 & 213.2 \tabularnewline \bottomrule
		\end{tabular}}
	}
        \vspace{-2mm}
	\caption{Comparisons of the robustness and generalization. ViT-B/16 and Swin-B models are pre-trained on ImageNet-21K. 'IN' denotes ImageNet. Performance metrics include Top-1 accuracy (\%) for IN-1K, IN-A, and IN-R datasets, with higher values ($\uparrow$) indicating better performance, and mean Corruption Error (mCE) for IN-C, with lower values ($\downarrow$) indicating better performance.}
	\label{table:ood}
\end{table*}

\subsection{Partial Fine-Tuning can Serve As A New Dimension for Model Soups}
Model Soups~\cite{wortsman2022model} improves the model performance and generalization ability by averaging weights fine-tuned with different configures. Its subsequent variations~\cite{rame2022diverse,rame2023model} focus on designing different training configures to diversify the averaged weights for better results. Although the training configures may differ, all the weights used in these methods are fully fine-tuned. In this paper, we claim and show that partial fine-tuning can serve as a new dimension for Model Soups, which provides a fresh perspective for future research in this direction.


\section{Experiments}

\subsection{Experimental Settings}
\label{subsec:exp_setting}
\textbf{Datasets}. We evaluate our method on a variety of datasets, which can be categorized into: \textit{1) General Classification Datasets}. To assess the effectiveness of FAPFT, we conduct experiments on commonly used image classification datasets, namely CIFAR-100~\cite{krizhevsky2009learning} and ImageNet-1K~\cite{deng2009imagenet}. 
\textit{2) FGVC Datasets}. We further extend our method to five Fine-Grained Visual Classification (FGVC) datasets to showcase its advantages, including NABirds~\cite{van2015building}, CUB-200-2011~\cite{wah2011caltech}, Oxford Flowers~\cite{nilsback2008automated}, Stanford Dogs~\cite{khosla2011novel}, and Stanford Cars~\cite{gebru2017fine}.

\noindent
\textbf{Models.} To show the generalization ability of our method, we conduct experiments on four different backbones: plain transformer-based ViT/B-16~\cite{dosovitskiy2020image}, hierarchical transformer-based Swin-B~\cite{liu2021swin}, CNN-structured ConvNeXt-B~\cite{liu2022convnet}, and MLP-structured AS-MLP-B~\cite{lian2021mlp}. Note that AS-MLP-B is pre-trained on ImageNet-1K while the others are pre-trained on ImageNet-21K.

\noindent
\textbf{Baselines}. In this study, we benchmark our approach with two basic fine-tuning methods as well as three parameter-efficient solutions:
1) Full fine-tuning, which updates all model parameters.
2) Linear probing, which only updates the classifier's parameters.
3) Adapter~\cite{hu2023llm}, which only updates parameters in inserted low-rank FFN modules.
4) VPT~\cite{jia2022visual}, which only updates newly incorporated prompt tokens.
5) SSF~\cite{lian2022scaling}, which only updates linear transformation parameters for modulating features.

\noindent
\textbf{Implementation Details}.
For general datasets, in line with SSF~\cite{lian2022scaling}, we train for 100 epochs with a 10 epoch warm-up on CIFAR-100, 30 epochs with a 5 epoch warm-up on ImageNet-1K, and employ the same strong data augmentation strategy. For five FGVC datasets, following VPT~\cite{jia2022visual}, we train for 100 epochs with a 10 epoch warm up on each FGVC dataset and adopt the same standard augmentations. 
For all experiments, we utilize the AdamW optimizer and the cosine decay learning rate scheduler.

In our partial fine-tuning strategies, pre-trained weights are selectively frozen at the layer level, where a layer refers to an entire residual-connected unit. For instance, a ViT's attention layer comprises a multi-head attention module and its preceding normalization layer, as depicted in Fig.~\ref{fig:main_comparisons}. Further, we define layers with identical structures and parameter counts as the homogeneous group. For example, in the ViT model, all FFN layers form a single homogeneous group, while in the Swin model, homogeneous FFN layers are segregated by stage. More details can be found in the Appendix.

For FAPFT deployment, we start with a fully fine-tuned model tailored to the specific dataset and architecture. By comparing this model to its pre-trained counterpart, we compute fine-tuned angles and subsequently organize layers into their respective homogeneous groups. Within each group, we select the top k layers with the largest or smallest angles for targeted partial fine-tuning. 

Notably, across all datasets and model architectures, each of our partial fine-tuning methods utilizes the same hyper-parameters as those applied in full fine-tuning and defaults to freezing non-residual modules.

\subsection{Experiments on Image Classification}

We conduct extensive experiments across a wide range of datasets, including CIFAR-100, ImageNet-1K, and various FGVC datasets, utilizing various architectures such as ViT, Swin, ConvNeXt, and AS-MLP. The results, detailed in Tab.~\ref{table:imagenet} and Tab.~\ref{table:fgvc}, consistently verify that Fine-tuned Angle guided Partial Fine-Tuning (FAPFT) surpasses in achieving both high accuracy and improved parameter efficiency.

Tab.~\ref{table:imagenet} shows the effectiveness of FAPFT on general datasets. As we can see, PEFT methods fall short of surpassing the baseline of full fine-tuning on the ImageNet-1K dataset. As a comparison, the hand-designed partial fine-tuning strategies of ATTN-Only and FFN-Only demonstrate comparable or even better performance with fewer parameters than full fine-tuning. Further, the proposed FAPFT approach achieves even better results, for the ViT model, even surpasses the baseline method by a significant margin. Moreover, FAPFT possesses greater universality and also works for other architectures like ConvNeXt-B and AS-MLP-B.
Additionally, on the CIFAR-100 dataset, FAPFT also outperforms other approaches consistently, confirming its robustness with different datasets.

As Tab.~\ref{table:fgvc} indicates, all three partial fine-tuning strategies (ATTN-Only, FFN-Only, and FAPFT) perform well on the five FGVC datasets, and averaged results highlight the benefits of partial fine-tuning than full fine-tuning. Although hand-designed ATTN-Only and FFN-Only methods can achieve good performance, the proposed FAPFT method stands out as the most effective, delivering a significant improvement of 90.17\% compared to 88.64\% with remarkably fewer parameters (30.69M VS 85.98M).

\begin{figure*}[h]
  \centering
  \subfloat[Training Latency]{
  \begin{subfigure}[b]{0.245\textwidth}
    \includegraphics[width=0.97\textwidth]{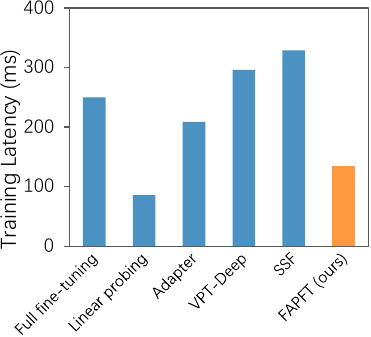}
  \end{subfigure}}
  \subfloat[Training Memory]{
  \begin{subfigure}[b]{0.245\textwidth}
    \includegraphics[width=0.97\textwidth]{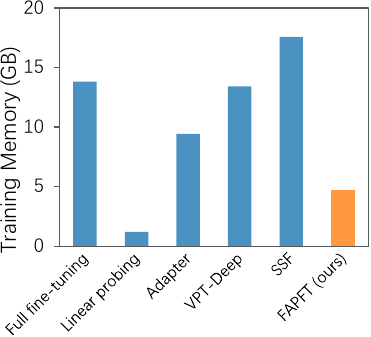}
  \end{subfigure}}
  \subfloat[Test Latency]{
  \begin{subfigure}[b]{0.245\textwidth}
    \includegraphics[width=0.97\textwidth]{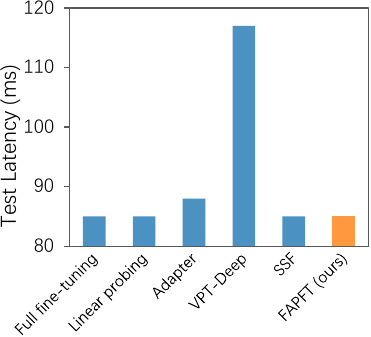}
  \end{subfigure}}
  \subfloat[Test Memory]{
  \begin{subfigure}[b]{0.245\textwidth}
    \includegraphics[width=0.97\textwidth]{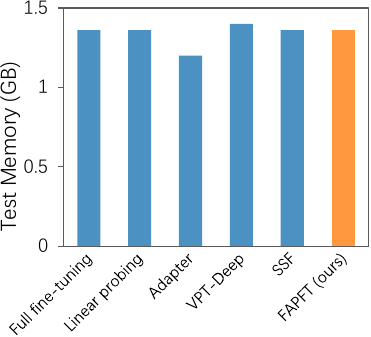}
  \end{subfigure}}
  \vspace{-3mm}
  \caption{Comprehensive comparisons of the computational cost among various fine-tuning methods, including (a) Training Latency, (b) Training Memory, (C) Test Latency, and (d) Test Memory.}
  \label{fig:cost}
\end{figure*}

\begin{figure}[t]
  \centering
  \subfloat[ViT-B/16 on CIFAR-100]{
  \begin{subfigure}{0.235\textwidth}
    \includegraphics[width=1.0\textwidth]{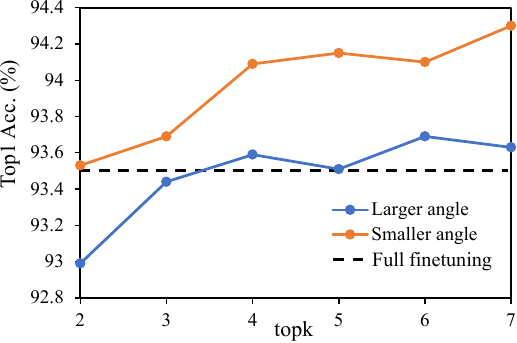}
    \label{subfig:ablation_cifar100}
  \end{subfigure}}
  \subfloat[ViT-B/16 on ImageNet-1K]{
  \begin{subfigure}{0.235\textwidth}
    \includegraphics[width=1.0\textwidth]{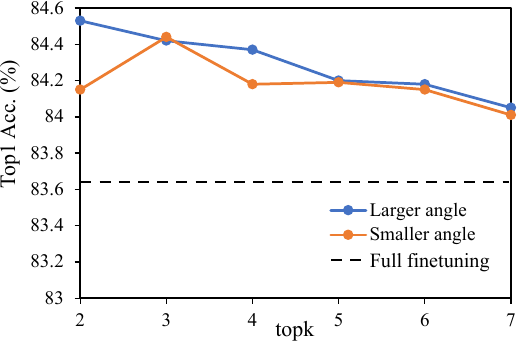}
    \label{subfig:ablation_imagenet}
  \end{subfigure}}
  \vspace{-2mm}
  \caption{Impact of two key components of our FAPFT, e.g., the magnitude of the fine-tuned angle for layer selection (larger or smaller angle), and the number of layers to be fine-tuned (topk). Subfigure (a) and (b) denote the results on the CIFAR-100 and ImageNet-1K datasets respectively. Both use the ViT-B/16 model pre-trained on ImageNet-21K. The dashed line represents the baseline performance of full fine-tuning (FFT). 
  }
  \label{fig:ablation}
\end{figure}

\subsection{Experiments on Model Soups}
In this section, we show how the proposed FAPFT approach can serve as a new dimension for Model Soups.
We conduct experiments using ViT-B/16 and Swin-B models on ImageNet-1K and CIFAR-100 datasets to showcase the superiority of FAPFT over full fine-tuning (FFT) in the model soups context. Specifically, we conduct five runs for each method with distinct configures, concluding with an aggregated soup. FFT's experiments vary in hyper-parameters, such as learning rate and seed. For FAPFT, we maintain consistent hyper-parameters across experiments, only altering the number of layers to be fine-tuned.

As shown in Tab.~\ref{table:model-soup}, individual runs highlight FAPFT's consistent outperformance in both model accuracy and parameter efficiency, ultimately resulting in Soup-FAPFT surpassing Soup-FFT in both metrics. Notably, compared to Soup-FFT based on the ViT-B/16 model on ImageNet-1K and CIFAR-100 datasets, Soup-FAPFT achieves an accuracy gain of 0.81\% and 0.32\%, alongside a significant parameter reduction of 322.7 and 283.2 million, respectively.  
Further, Soup-FAPFT consistently outperforms Soup-FFT for the Swin-B model on different datasets while maintaining parameter efficiency. These results indicate the potential of FAPFT to not only strengthen individual fine-tuned models but also uplift the collective performance for the Model Soups framework in a parameter-efficient manner.

\subsection{Experiments on Robustness and OOD Datasets}
We further evaluate the robustness and Out-Of-Distribution (OOD) capabilities of the proposed FAPFT method on ImageNet-A, ImageNet-R, and ImageNet-C datasets. All models are fine-tuned on ImageNet-1K. Results are listed in Tab.~\ref{table:ood}.
In most scenarios, partial fine-tuning methods (namely ATTN-Only, FFN-Only, and our FAPFT) always yield models with better robustness and OOD generalization compared to full fine-tuning. For example, ViT-B/16 finetuned via FAPFT shows a 0.91\% increase in IID accuracy on ImageNet-1K as well as notable gains of 7.43\%, 1.29\%, and 1.99\% on ImageNet-A/R/C respectively. These gains likely stem from FAPFT’s uniqueness in fine-tuning only the most influential layers, ensuring the most effective adaptation for challenging tasks.
Furthermore, when combined with Model Soups, the resultant averaged model consistently outperforms its individual counterpart across all metrics. What's more, the FAPFT-based soup achieves comprehensive improvements with reduced computation cost over the full fine-tuning (FFT) based soup, across various model architectures and datasets.

\subsection{Ablation Studies}
In this section, we explore two critical components in the proposed FAPFT strategy, e.g., the magnitude of the fine-tuned angle for layer selection (larger or smaller angle) and the number of layers to be fine-tuned (top-k). We conduct experiments using the ViT-B/16 model on CIFAR-100 and ImageNet-1K datasets to gain insights into their impact on performance.
Results are presented in Fig.~\ref{fig:ablation}. For the easier dataset of CIFAR-100, fine-tuning layers with smaller fine-tuned angles consistently leads to significantly better performance than fine-tuning layers with larger fine-tuned angles. Conversely, on the more complex dataset of ImageNet-1K, fine-tuning layers with larger fine-tuned angles is more advantageous, which performs better across multiple top-k values and attains a notable peak performance of 84.53\% in the setting of top2 larger angles. 
These observations further confirm the idea of our FAPFT method:
partial finetuning based on smaller fine-tuned angles benefits the simpler datasets or tasks since it can maximally maintain the pre-trained knowledge, whereas partial finetuning based on larger angles benefits the more complex datasets or tasks since it can realize the most effective adaptation.

\subsection{Computational cost}
To evaluate the efficiency of the proposed FAPFT, we compare the computational cost of various fine-tuning methods in Fig.~\ref{fig:cost}. We conduct the training and inference stages using a batch size of 128 and employ mixed precision training. All measurements are obtained using a single NVIDIA A100 GPU with 80GB of GPU memory. To ensure a fair comparison, we utilize identical settings to those used for ViT-B/16 on ImageNet-1K in Tab.~\ref{table:imagenet}. Specifically, we employ a reduction factor of 64 for the Adapter and 50 prompts for VPT-Deep. 
As we can see, most methods require additional cost. For example, Adapter increases the test latency, VPT-Deep increases the training latency, test latency, and test memory, while SSF increases the training latency and training memory. As a comparison, FAPFT exhibits much lower training latency and memory, and the same test latency and memory, demonstrating its great potential for future research and practical applications. 

\section{Conclusion}
In this paper, we present partial fine-tuning as an innovative and promising approach capable of concurrently improving performance and parameter efficiency. We first find that partial fine-tuning of specific functional layers can achieve better performance with fewer tuned parameters than full fine-tuning, and selecting appropriate layers has a substantial impact on partial fine-tuning. Further, we propose a general partial fine-tuning method via a novel fine-tuned angle metric, adaptively selecting more appropriate layers in various scenarios.
Extensive experiments across diverse datasets and architectures validate the substantial potential of partial fine-tuning.
{
    \small
    \bibliographystyle{ieeenat_fullname}
    \bibliography{main}

\begin{thebibliography}{38}
\providecommand{\natexlab}[1]{#1}
\providecommand{\url}[1]{\texttt{#1}}
\expandafter\ifx\csname urlstyle\endcsname\relax
  \providecommand{\doi}[1]{doi: #1}\else
  \providecommand{\doi}{doi: \begingroup \urlstyle{rm}\Url}\fi

\bibitem[Arora et~al.(2018)Arora, Li, and Lyu]{arora2018theoretical}
Sanjeev Arora, Zhiyuan Li, and Kaifeng Lyu.
\newblock Theoretical analysis of auto rate-tuning by batch normalization.
\newblock \emph{arXiv preprint arXiv:1812.03981}, 2018.

\bibitem[Carbonnelle and De~Vleeschouwer(2019)]{carbonnelle2019layer}
Simon Carbonnelle and Christophe De~Vleeschouwer.
\newblock Layer rotation: a surprisingly simple indicator of generalization in deep networks?
\newblock In \emph{ICML 2019 Workshop on Identifying and Understanding Deep Learning Phenomena}, 2019.

\bibitem[Deng et~al.(2009)Deng, Dong, Socher, Li, Li, and Fei-Fei]{deng2009imagenet}
Jia Deng, Wei Dong, Richard Socher, Li-Jia Li, Kai Li, and Li Fei-Fei.
\newblock Imagenet: A large-scale hierarchical image database.
\newblock In \emph{2009 IEEE conference on computer vision and pattern recognition}, pages 248--255. Ieee, 2009.

\bibitem[Devlin et~al.(2018)Devlin, Chang, Lee, and Toutanova]{devlin2018bert}
Jacob Devlin, Ming-Wei Chang, Kenton Lee, and Kristina Toutanova.
\newblock Bert: Pre-training of deep bidirectional transformers for language understanding.
\newblock \emph{arXiv preprint arXiv:1810.04805}, 2018.

\bibitem[Dong et~al.(2023)Dong, Yan, Lin, and Wang]{dong2023efficient}
Wei Dong, Dawei Yan, Zhijun Lin, and Peng Wang.
\newblock Efficient adaptation of large vision transformer via adapter re-composing.
\newblock \emph{arXiv preprint arXiv:2310.06234}, 2023.

\bibitem[Dosovitskiy et~al.(2020)Dosovitskiy, Beyer, Kolesnikov, Weissenborn, Zhai, Unterthiner, Dehghani, Minderer, Heigold, Gelly, et~al.]{dosovitskiy2020image}
Alexey Dosovitskiy, Lucas Beyer, Alexander Kolesnikov, Dirk Weissenborn, Xiaohua Zhai, Thomas Unterthiner, Mostafa Dehghani, Matthias Minderer, Georg Heigold, Sylvain Gelly, et~al.
\newblock An image is worth 16x16 words: Transformers for image recognition at scale.
\newblock \emph{arXiv preprint arXiv:2010.11929}, 2020.

\bibitem[Gebru et~al.(2017)Gebru, Krause, Wang, Chen, Deng, and Fei-Fei]{gebru2017fine}
Timnit Gebru, Jonathan Krause, Yilun Wang, Duyun Chen, Jia Deng, and Li Fei-Fei.
\newblock Fine-grained car detection for visual census estimation.
\newblock In \emph{Proceedings of the AAAI Conference on Artificial Intelligence}, 2017.

\bibitem[He et~al.(2022)He, Chen, Xie, Li, Doll{\'a}r, and Girshick]{he2022masked}
Kaiming He, Xinlei Chen, Saining Xie, Yanghao Li, Piotr Doll{\'a}r, and Ross Girshick.
\newblock Masked autoencoders are scalable vision learners.
\newblock In \emph{Proceedings of the IEEE/CVF conference on computer vision and pattern recognition}, pages 16000--16009, 2022.

\bibitem[Houlsby et~al.(2019)Houlsby, Giurgiu, Jastrzebski, Morrone, De~Laroussilhe, Gesmundo, Attariyan, and Gelly]{houlsby2019parameter}
Neil Houlsby, Andrei Giurgiu, Stanislaw Jastrzebski, Bruna Morrone, Quentin De~Laroussilhe, Andrea Gesmundo, Mona Attariyan, and Sylvain Gelly.
\newblock Parameter-efficient transfer learning for nlp.
\newblock In \emph{International Conference on Machine Learning}, pages 2790--2799. PMLR, 2019.

\bibitem[Hu et~al.(2021)Hu, Shen, Wallis, Allen-Zhu, Li, Wang, Wang, and Chen]{hu2021lora}
Edward~J Hu, Yelong Shen, Phillip Wallis, Zeyuan Allen-Zhu, Yuanzhi Li, Shean Wang, Lu Wang, and Weizhu Chen.
\newblock Lora: Low-rank adaptation of large language models.
\newblock \emph{arXiv preprint arXiv:2106.09685}, 2021.

\bibitem[Hu et~al.(2020)Hu, Liang, Guo, Wan, Zhang, Wei, Gu, and Sun]{hu2020angle}
Yiming Hu, Yuding Liang, Zichao Guo, Ruosi Wan, Xiangyu Zhang, Yichen Wei, Qingyi Gu, and Jian Sun.
\newblock Angle-based search space shrinking for neural architecture search.
\newblock In \emph{Computer Vision--ECCV 2020: 16th European Conference, Glasgow, UK, August 23--28, 2020, Proceedings, Part XIX 16}, pages 119--134. Springer, 2020.

\bibitem[Hu et~al.(2023)Hu, Lan, Wang, Xu, Lim, Lee, Bing, and Poria]{hu2023llm}
Zhiqiang Hu, Yihuai Lan, Lei Wang, Wanyu Xu, Ee-Peng Lim, Roy Ka-Wei Lee, Lidong Bing, and Soujanya Poria.
\newblock Llm-adapters: An adapter family for parameter-efficient fine-tuning of large language models.
\newblock \emph{arXiv preprint arXiv:2304.01933}, 2023.

\bibitem[Huang et~al.(2023)Huang, Liu, Lin, Pang, Du, and Lin]{huang2023lorahub}
Chengsong Huang, Qian Liu, Bill~Yuchen Lin, Tianyu Pang, Chao Du, and Min Lin.
\newblock Lorahub: Efficient cross-task generalization via dynamic lora composition.
\newblock \emph{arXiv preprint arXiv:2307.13269}, 2023.

\bibitem[Ilharco et~al.(2022)Ilharco, Ribeiro, Wortsman, Gururangan, Schmidt, Hajishirzi, and Farhadi]{ilharco2022editing}
Gabriel Ilharco, Marco~Tulio Ribeiro, Mitchell Wortsman, Suchin Gururangan, Ludwig Schmidt, Hannaneh Hajishirzi, and Ali Farhadi.
\newblock Editing models with task arithmetic.
\newblock \emph{arXiv preprint arXiv:2212.04089}, 2022.

\bibitem[Jia et~al.(2022)Jia, Tang, Chen, Cardie, Belongie, Hariharan, and Lim]{jia2022visual}
Menglin Jia, Luming Tang, Bor-Chun Chen, Claire Cardie, Serge Belongie, Bharath Hariharan, and Ser-Nam Lim.
\newblock Visual prompt tuning.
\newblock In \emph{European Conference on Computer Vision}, pages 709--727. Springer, 2022.

\bibitem[Khosla et~al.(2011)Khosla, Jayadevaprakash, Yao, and Li]{khosla2011novel}
Aditya Khosla, Nityananda Jayadevaprakash, Bangpeng Yao, and Fei-Fei Li.
\newblock Novel dataset for fine-grained image categorization: Stanford dogs.
\newblock In \emph{Proc. CVPR workshop on fine-grained visual categorization (FGVC)}. Citeseer, 2011.

\bibitem[Krizhevsky et~al.(2009)Krizhevsky, Hinton, et~al.]{krizhevsky2009learning}
Alex Krizhevsky, Geoffrey Hinton, et~al.
\newblock Learning multiple layers of features from tiny images.
\newblock 2009.

\bibitem[Li and Liang(2021)]{li2021prefix}
Xiang~Lisa Li and Percy Liang.
\newblock Prefix-tuning: Optimizing continuous prompts for generation.
\newblock \emph{arXiv preprint arXiv:2101.00190}, 2021.

\bibitem[Li and Arora(2019)]{li2019exponential}
Zhiyuan Li and Sanjeev Arora.
\newblock An exponential learning rate schedule for deep learning.
\newblock \emph{arXiv preprint arXiv:1910.07454}, 2019.

\bibitem[Lian et~al.(2021)Lian, Yu, Sun, and Gao]{lian2021mlp}
Dongze Lian, Zehao Yu, Xing Sun, and Shenghua Gao.
\newblock As-mlp: An axial shifted mlp architecture for vision.
\newblock \emph{arXiv preprint arXiv:2107.08391}, 2021.

\bibitem[Lian et~al.(2022)Lian, Zhou, Feng, and Wang]{lian2022scaling}
Dongze Lian, Daquan Zhou, Jiashi Feng, and Xinchao Wang.
\newblock Scaling \& shifting your features: A new baseline for efficient model tuning.
\newblock \emph{Advances in Neural Information Processing Systems}, 35:\penalty0 109--123, 2022.

\bibitem[Liang et~al.(2023)Liang, Bai, Qiao, Ren, Sun, Ye, Yan, Ma, Zuo, and Ouyang]{liang2023rethinking}
Chaoqi Liang, Weiqiang Bai, Lifeng Qiao, Yuchen Ren, Jianle Sun, Peng Ye, Hongliang Yan, Xinzhu Ma, Wangmeng Zuo, and Wanli Ouyang.
\newblock Rethinking the bert-like pretraining for dna sequences.
\newblock \emph{arXiv preprint arXiv:2310.07644}, 2023.

\bibitem[Liu et~al.(2021)Liu, Lin, Cao, Hu, Wei, Zhang, Lin, and Guo]{liu2021swin}
Ze Liu, Yutong Lin, Yue Cao, Han Hu, Yixuan Wei, Zheng Zhang, Stephen Lin, and Baining Guo.
\newblock Swin transformer: Hierarchical vision transformer using shifted windows.
\newblock In \emph{Proceedings of the IEEE/CVF international conference on computer vision}, pages 10012--10022, 2021.

\bibitem[Liu et~al.(2022)Liu, Mao, Wu, Feichtenhofer, Darrell, and Xie]{liu2022convnet}
Zhuang Liu, Hanzi Mao, Chao-Yuan Wu, Christoph Feichtenhofer, Trevor Darrell, and Saining Xie.
\newblock A convnet for the 2020s.
\newblock In \emph{Proceedings of the IEEE/CVF conference on computer vision and pattern recognition}, pages 11976--11986, 2022.

\bibitem[Nilsback and Zisserman(2008)]{nilsback2008automated}
Maria-Elena Nilsback and Andrew Zisserman.
\newblock Automated flower classification over a large number of classes.
\newblock In \emph{2008 Sixth Indian conference on computer vision, graphics \& image processing}, pages 722--729. IEEE, 2008.

\bibitem[Rame et~al.(2022)Rame, Kirchmeyer, Rahier, Rakotomamonjy, Gallinari, and Cord]{rame2022diverse}
Alexandre Rame, Matthieu Kirchmeyer, Thibaud Rahier, Alain Rakotomamonjy, Patrick Gallinari, and Matthieu Cord.
\newblock Diverse weight averaging for out-of-distribution generalization.
\newblock \emph{Advances in Neural Information Processing Systems}, 35:\penalty0 10821--10836, 2022.

\bibitem[Rame et~al.(2023)Rame, Ahuja, Zhang, Cord, Bottou, and Lopez-Paz]{rame2023model}
Alexandre Rame, Kartik Ahuja, Jianyu Zhang, Matthieu Cord, L{\'e}on Bottou, and David Lopez-Paz.
\newblock Model ratatouille: Recycling diverse models for out-of-distribution generalization.
\newblock 2023.

\bibitem[Shen et~al.(2021)Shen, Liu, Qin, Savvides, and Cheng]{shen2021partial}
Zhiqiang Shen, Zechun Liu, Jie Qin, Marios Savvides, and Kwang-Ting Cheng.
\newblock Partial is better than all: revisiting fine-tuning strategy for few-shot learning.
\newblock In \emph{Proceedings of the AAAI Conference on Artificial Intelligence}, pages 9594--9602, 2021.

\bibitem[Tang et~al.(2023)Tang, Ye, Li, Lin, Chen, He, Yu, and Ouyang]{tang2023boosting}
Shengji Tang, Peng Ye, Baopu Li, Weihao Lin, Tao Chen, Tong He, Chong Yu, and Wanli Ouyang.
\newblock Boosting residual networks with group knowledge.
\newblock \emph{arXiv preprint arXiv:2308.13772}, 2023.

\bibitem[Touvron et~al.(2022)Touvron, Cord, El-Nouby, Verbeek, and J{\'e}gou]{touvron2022three}
Hugo Touvron, Matthieu Cord, Alaaeldin El-Nouby, Jakob Verbeek, and Herv{\'e} J{\'e}gou.
\newblock Three things everyone should know about vision transformers.
\newblock In \emph{European Conference on Computer Vision}, pages 497--515. Springer, 2022.

\bibitem[Van~Horn et~al.(2015)Van~Horn, Branson, Farrell, Haber, Barry, Ipeirotis, Perona, and Belongie]{van2015building}
Grant Van~Horn, Steve Branson, Ryan Farrell, Scott Haber, Jessie Barry, Panos Ipeirotis, Pietro Perona, and Serge Belongie.
\newblock Building a bird recognition app and large scale dataset with citizen scientists: The fine print in fine-grained dataset collection.
\newblock In \emph{Proceedings of the IEEE conference on computer vision and pattern recognition}, pages 595--604, 2015.

\bibitem[Wah et~al.(2011)Wah, Branson, Welinder, Perona, and Belongie]{wah2011caltech}
Catherine Wah, Steve Branson, Peter Welinder, Pietro Perona, and Serge Belongie.
\newblock The caltech-ucsd birds-200-2011 dataset.
\newblock 2011.

\bibitem[Wortsman et~al.(2022)Wortsman, Ilharco, Gadre, Roelofs, Gontijo-Lopes, Morcos, Namkoong, Farhadi, Carmon, Kornblith, et~al.]{wortsman2022model}
Mitchell Wortsman, Gabriel Ilharco, Samir~Ya Gadre, Rebecca Roelofs, Raphael Gontijo-Lopes, Ari~S Morcos, Hongseok Namkoong, Ali Farhadi, Yair Carmon, Simon Kornblith, et~al.
\newblock Model soups: averaging weights of multiple fine-tuned models improves accuracy without increasing inference time.
\newblock In \emph{International Conference on Machine Learning}, pages 23965--23998. PMLR, 2022.

\bibitem[Yadav et~al.(2023)Yadav, Tam, Choshen, Raffel, and Bansal]{yadav2023resolving}
Prateek Yadav, Derek Tam, Leshem Choshen, Colin Raffel, and Mohit Bansal.
\newblock Resolving interference when merging models.
\newblock \emph{arXiv preprint arXiv:2306.01708}, 2023.

\bibitem[Yang et~al.(2023)Yang, Yang, Jin, and Chen]{yang2023revisiting}
Taojiannan Yang, Linjie Yang, Xiaojie Jin, and Chen Chen.
\newblock Revisiting training-free nas metrics: An efficient training-based method.
\newblock In \emph{Proceedings of the IEEE/CVF Winter Conference on Applications of Computer Vision}, pages 4751--4760, 2023.

\bibitem[Ye et~al.(2022)Ye, Tang, Li, Chen, and Ouyang]{ye2022stimulative}
Peng Ye, Shengji Tang, Baopu Li, Tao Chen, and Wanli Ouyang.
\newblock Stimulative training of residual networks: A social psychology perspective of loafing.
\newblock \emph{Advances in Neural Information Processing Systems}, 35:\penalty0 3596--3608, 2022.

\bibitem[Ye et~al.(2023)Ye, He, Tang, Li, Chen, Bai, and Ouyang]{ye2023stimulative}
Peng Ye, Tong He, Shengji Tang, Baopu Li, Tao Chen, Lei Bai, and Wanli Ouyang.
\newblock Stimulative training++: Go beyond the performance limits of residual networks.
\newblock \emph{arXiv preprint arXiv:2305.02507}, 2023.

\bibitem[Zhang et~al.(2021)Zhang, Hou, Zhang, and Sun]{zhang2021neural}
Xuanyang Zhang, Pengfei Hou, Xiangyu Zhang, and Jian Sun.
\newblock Neural architecture search with random labels.
\newblock In \emph{Proceedings of the IEEE/CVF conference on computer vision and pattern recognition}, pages 10907--10916, 2021.

\end{thebibliography}
}

\appendix
\clearpage
\setcounter{page}{1}
\maketitlesupplementary

\begin{table*}[t]
    \centering
    \scalebox{1.05}{
        \begin{tabular}{ l  l  l  l  l  l }
            \toprule
            Dataset & Description & Classes & Train size & Val size & Test size \\ 
            \midrule
            \multicolumn{2}{l}{General Classification Datasets} \\
            \cmidrule{1-6}
            CIFAR-100 & \multirow{2}{*}{General image classification} & 100 &50,000 & - &10,000  \\
            ImageNet-1K &  & 1,000 & 1,281,167  & 50,000 & 150,000 \\
            \midrule
            \multicolumn{2}{l}{Fine-Grained Visual Classification (FGVC)}  \\
            \cmidrule{1-6}
            CUB-200-2011 & Fine-grained bird species recognition &200 &5,394$^{\star}$	&600$^{\star}$ &5,794	\\
            NABirds & Fine-grained bird species recognition &55 &21,536$^{\star}$	&2,393$^{\star}$	&24,633	\\
            Oxford Flowers & Fine-grained flower species recognition &102 &1,020	&1,020	&6,149  \\
            Stanford Dogs &Fine-grained dog species recognition  &120  &10,800$^{\star}$	&1,200$^{\star}$ &8,580 \\
            Stanford Cars & Fine-grained car classification  &196   &7,329$^{\star}$	&815$^{\star}$	&8,041 \\
            \midrule
            \multicolumn{2}{l}{Robustness and Out-of-Distribution Dataset} \\
            \cmidrule{1-6}
            ImageNet-A & \multirow{3}{*}{Robustness \& OOD}  & 200  & \multicolumn{3}{c}{7,500}  \\
            ImageNet-R &  & 200 &  \multicolumn{3}{c}{30,000} \\
            ImageNet-C &   & 1,000&  \multicolumn{3}{c}{75 $\times$ 50,000} \\
            \bottomrule
        \end{tabular}
    }
    \vspace{-2mm}
    \caption{The detailed specifications of kinds of datasets. This table is partially borrowed from SSF~\cite{lian2022scaling}. $^{\star}$ denotes that we employ the random train/val split strategy following VPT~\cite{jia2022visual}.}
    \label{table:datasets}
\end{table*}

\section{Detailed Descriptions for Datasets}
We present the detailed specifications of datasets used for evaluation in Tab.~\ref{table:datasets}, including the number of classes, the size of the train set, val set and test set respectively.

\section{Implementation Details of FAPFT}
Our Fine-tuned Angle guided Partial Fine-Tuning (FAPFT) introduces a novel fine-tuned angle metric, which computes the angle between the pre-trained and fine-tuned weights of a specific layer in a model, to guide the selection of appropriate layers for partial fine-tuning. This section provides the implementation specifics of FAPFT.

\subsection{Homogeneous Group of different architectures}
As briefly described in Sec.~\ref{subsec:exp_setting} of the manuscript, the proposed FAPFT approach involves freezing weights at the layer level, where a layer denotes an entire residual unit. Further, we categorize layers with identical structures and parameter counts into homogeneous groups. Detailed categorizations of layers and homogeneous groups for different evaluated model architectures, namely ViT-B/16, Swin-B, ConvNeXt-B and AS-MLP-B, are provided in Tab.~\ref{table:homogeneous_group}.

The one-stage ViT-B/16 is structured with 12 identical blocks, comprising two layer categories: ATTN layer and FFN layer, each consisting of a preceding LayerNorm module and a specific module (MSA for the former, FFN for the latter). These layers can be further organized into two homogeneous groups: all ATTN layers and all FFN layers.

The multi-stage Swin-B model consists of four stages with 2, 2, 18, and 2 blocks in each respective stage. Similar to ViT-B/16, Swin-B contains only two types of layers: FFN and ATTN. However, it has eight homogeneous groups due to the distinct stages. All ATTN or FFN layers within each stage share the same structure and parameter count, resulting in each stage being grouped separately. AS-MLP-B follows a similar categorization schema to that of Swin-B.

The ConvNeXt-B has only one type of residual unit in its architecture: the basic block. Therefore, all blocks within each stage form a single homogeneous group.

\begin{table*}[t]
    \centering
    \renewcommand\arraystretch{1.5}
    \scalebox{0.95}{
        \begin{tabular}{c  c  c  c}
            \toprule
            Model & Number of Blocks & Layer Category & Homogeneous Group Category \\ \hline
            \multirow{2}{*}{ViT-B/16} & \multirow{2}{*}{12} & ATTN layer (LayerNorm + MSA module) & All ATTN layers \\
            & & FFN layer (LayerNorm + FFN module) & All FFN layers \\ \hline
            \multirow{2}{*}{Swin-B} & \multirow{2}{*}{[2, 2, 18, 2]} & ATTN layer (LayerNorm + MSA module) & All ATTN layers within each stage \\
                                    & & FFN layer (LayerNorm + FFN module) & All FFN layers within each stage \\ \hline
            \multirow{2}{*}{AS-MLP-B} & \multirow{2}{*}{[2, 2, 18, 2]} & AS layer (LayerNorm + AS module) & All AS layers within each stage \\
                                      & & MLP layer (LayerNorm + MLP module) & All MLP layers within each stage \\ \hline
            ConvNeXt-B & [3, 3, 27, 3] & ConvNeXt block & All ConvNeXt blocks within each stage\\ \hline
        \end{tabular}
    }
    \vspace{-2mm}
    \caption{Configures of various model architectures, including number of blocks, categories of layers and homogeneous groups.}
    \label{table:homogeneous_group}
\end{table*}

\begin{table*}[t]
    \centering
    \renewcommand\arraystretch{1.5}
    \scalebox{1.0}{
        \begin{tabular}{c|c|c|c|c|c}
            \toprule
            \multirow{2}{*}{Model} & \multirow{2}{*}{Number of Blocks} & \multicolumn{2}{c|}{Easy Dataset} & \multicolumn{2}{c}{Challenging Dataset} \\ \cline{3-6}
             & & default topk & suggested topk range & default topk & suggested topk range \\ \hline
            ViT-B/16 & 12 & 4 & 2-7 & 4 & 2-7 \\ 
            Swin-B & {[}2, 2, 18, 2{]} & {[}2, 2, 6, 1{]} & {[}2, 2, 4-8, 1-2{]} & {[}0, 0, 6, 2{]} & {[}0, 0, 4-8, 1-2{]} \\ 
            ConvNeXt-B & {[}3, 3, 27, 3{]} & {[}3, 3, 9, 1{]} & {[}3, 3, 7-11, 1-3{]} & {[}0, 0, 9, 3{]} & {[}0, 0, 7-11, 1-3{]} \\ 
            AS-MLP-B & {[}2, 2, 18, 2{]} & {[}2, 2, 6, 1{]} & {[}2, 2, 4-8, 1-2{]} & - & - \\ \hline
        \end{tabular}
    }
    \vspace{-2mm}
    \caption{Guidelines for tuning topk across different architectures and datasets, where topk denotes the number of layers selected for fine-tuning within each homogeneous group of the corresponding stage. This table presents the recommended starting setting (default topk) and the adaptive range (suggested topk range) for the topk hyper-parameter when employing FAPFT across various models and datasets.}
    \label{table:guidelines}
\end{table*}

\subsection{FAPFT on different datasets and architectures}
Applying FAPFT across various datasets and model architectures requires tuning just two hyper-parameters: 1) the magnitude of fine-tuned angles for layer selection, which is adjusted between large angle and small angle, and determines only fine-tuning either layers with largest or smallest fine-tuned angles; 2) topk, which indicates the number of layers to be fine-tuned within the homogeneous group.

The magnitude of fine-tuned angles for layer selection is determined by the complexity of the dataset: small angle for easy tasks while large angle for complex tasks. For two general classification datasets utilized in our study, CIFAR-100 is easy while ImageNet-1K is challenging. For five FGVC datasets, StanfordDogs and OxfordFlowers are considered less complex, while the others are not. The reason is that, ViT-B/16 consistently achieves higher performance on both StanfordDogs and OxfordFlowers compared to the performance on other datasets, as shown in Tab.~\ref{table:fgvc} of the manuscript, confirming this categorization.

When applying FAPFT across different architectures, we assign each stage of the model its own topk, which determines the number of layers selected for fine-tuning within each homogeneous group of the corresponding stage. As shown in Tab~\ref{table:guidelines}, for one-stage models like ViT-B/16, one topk is sufficient, indicating the number of ATTN and FFN layers to be fine-tuned. For models with a four-stage structure, such as Swin-B and ConvNeXt-B, FAPFT defaults to fine-tuning the first two stages for easy datasets while freezing for complex datasets and focusing on tuning the topk of the last two stages. 

\subsection{Guidelines for FAPFT}
As shown in Tab.~\ref{table:guidelines}, we provide guidelines for tuning topk (the number of layers selected for fine-tuning within each homogeneous group of the corresponding stage) when employing FAPFT on various models and datasets. Empirical analysis indicates that the default topk setting are robust to achieve a promising performance. Further, to push the boundaries of model efficacy, practitioners are encouraged to explore within the suggested topk range for potential performance enhancements.

For instance, when utilizing our FAPFT method on the ViT-B/16, the topk setting of 4 is expected to yield promising outcomes. Further tuning of topk within the range of [2,7] may lead to even better performance.

For other four-stage model architectures, we simply adjust the topk of the last two stages. It is recommended to consistently fine-tune the first two stages for easy datasets, which is reflected in setting the topk value equal to the number of blocks. Conversely, when facing challenging datasets, consistently freezing the first two stages is better, as indicated by setting the topk value to zero. 

\subsection{Overall Procedure}
We present the procedure of FAPFT on a dataset $D$ with a model architecture $M$. Given the pre-trained weights $W_p$ and fine-tuning configure $C$, each step is detailed below:

\begin{enumerate}
    \item \textbf{Fully fine-tune the pre-trained model:} Utilizing the fine-tuning configure $C$, we fully fine-tune the model $M$ with the pre-trained weights $W_p$ on the dataset $D$, obtaining the fine-tuned weights $W_f$.
    
    \item \textbf{Determine FAPFT hyper-parameters:} 1) Magnitude of fine-tuned angles for layer selection. If the dataset $D$ is challenging, select layers with the largest fine-tuned angles within each homogeneous group for fine-tuning. Conversely, if the dataset is simple, opt for layers with the smallest fine-tuned angles. 2) Follow the guidelines to set topk (the number of layers to be fine-tuned within each homogeneous group of the corresponding stage) for the dataset $D$ and model architecture $M$.
    
    \item \textbf{Fine-tuned Angle guided layer selection:} 1) Following Eq.~\ref{eq:finetuned_angle} of the manuscript, calculate the fine-tuned angle for each layer. 2) Categorize layers into homogeneous groups. 3) Guided by FAPFT hyper-parameters, select layers for fine-tuning within each homogeneous group. 4) Finalize the list $L$ of layers to be fine-tuned.

    \item \textbf{Partially fine-tune the pre-trained model:} For the model $M$ equipped with the pre-trained weights $W_p$, freeze all layers and non-residual modules except those listed in $L$. Subsequently, fine-tune the partially frozen model on the dataset $D$ using the configure $C$.
\end{enumerate}

\section{Comprehensive results of tuning topk} 
In the ablation studies section, we present experiments on tuning the topk hyper-parameter when applying FAPFT to the ViT-B/16 model on ImageNet-1K and CIFAR-100 datasets. Moreover, we provide further results of tuning topk with Swin-B, ConvNeXt-B and AS-MLP-B models on these two datasets in Tab.~\ref{table:tuning_topk}. Notably, we fine-tune only layers with the largest fine-tuned angles for ImageNet-1K and layers with smallest fine-tuned angles for CIFAR-100.

\begin{table*}[t]
    \centering
    \hspace{15mm}
    \subfloat[Swin-B on ImageNet-1K]{
        \setlength{\tabcolsep}{12pt}
        \renewcommand\arraystretch{1.25}
        \scalebox{0.85}{
            \begin{tabular}{c|c c}
                \toprule
                topk & Acc & Params. \\ \midrule
                {[}2, 2, 18, 2{]}  & 85.07 & 87.77 \\ \midrule
                {[}0, 0, 3, 2{]} & 85.11 & 35.69 \\
                {[}0, 0, 4, 2{]} & 85.13 & 38.85 \\
                {[}0, 0, 5, 2{]} & 85.17 & 42.01 \\
                {[}0, 0, 6, 2{]} & 85.15 & 45.16 \\
                \bottomrule
            \end{tabular}
        }
    }
    \hspace{15mm}
    \subfloat[ConvNeXt-B on ImageNet-1K]{
        \setlength{\tabcolsep}{12pt}
        \renewcommand\arraystretch{1.25}
        \scalebox{0.85}{
            \begin{tabular}{c|c c}
                \toprule
                topk & Acc & Params. \\ \midrule
                {[}3, 3, 27, 3{]} & 85.49 & 88.59 \\ \midrule
                {[}0, 0, 6, 3{]} & 85.39 & 41.89 \\
                {[}0, 0, 7, 3{]} & 85.32 & 44.02 \\
                {[}0, 0, 8, 3{]} & 85.38 & 46.15 \\
                {[}0, 0, 9, 3{]} & 85.39 & 48.28 \\
                \bottomrule
            \end{tabular}
        }
    }
    \vspace{3mm}
    \newline
    \subfloat[Swin-B on CIFAR-100]{
        \setlength{\tabcolsep}{12pt}
        \renewcommand\arraystretch{1.25}
        \scalebox{0.82}{
            \begin{tabular}{c|c c}
                \toprule
                topk & Acc & Params. \\ \midrule
                {[}2, 2, 18, 2{]} & 93.77 & 86.85 \\ \midrule
                {[}1, 1, 5, 1{]} & 94.07 & 29.47 \\
                {[}1, 1, 6, 1{]} & 94.00 & 32.62 \\
                {[}2, 2, 5, 1{]} & 94.00 & 30.46 \\
                {[}2, 2, 6, 1{]} & 94.07 & 33.61 \\
                \bottomrule
            \end{tabular}
        }
    }
    \hspace{2mm}
    \subfloat[ConvNeXt-B on CIFAR-100]{
        \setlength{\tabcolsep}{12pt}
        \renewcommand\arraystretch{1.25}
        \scalebox{0.82}{
            \begin{tabular}{c|c c}
                \toprule
                topk & Acc & Params. \\ \midrule
                {[}3, 3, 27, 3{]} & 94.04 & 87.67 \\ \midrule
                {[}2, 2, 9, 1{]} & 93.75 & 31.81 \\
                {[}3, 3, 9, 1{]} & 93.90 & 32.49 \\
                {[}3, 3, 11, 1{]} & 93.95 & 36.74 \\
                {[}3, 3, 11, 2{]} & 94.05 & 45.19 \\
                \bottomrule
            \end{tabular}
        }
    }
    \hspace{2mm}
    \subfloat[AS-MLP-B on CIFAR-100]{
        \setlength{\tabcolsep}{12pt}
        \renewcommand\arraystretch{1.25}
        \scalebox{0.82}{
            \begin{tabular}{c|c c}
                \toprule
                topk & Acc & Params. \\ \midrule
                {[}2, 2, 18, 2{]} & 90.04 & 86.83 \\ \midrule
                {[}1, 1, 6, 1{]} & 90.48 & 32.62 \\
                {[}1, 1, 6, 2{]} & 90.53 & 45.22 \\
                {[}2, 2, 6, 1{]} & 90.64 & 33.61 \\
                {[}2, 2, 6, 2{]} & 90.74 & 46.21 \\
                \bottomrule
            \end{tabular}
        }
    }
\vspace{-2mm}
\caption{Results of tuning FAPFT's topk with Swin-B, ConvNeXt-B and AS-MLP-B models on ImageNet-1K and CIFAR-100 datasets respectively. The second row provides the performance and parameters of full fine-tuning.}
\label{table:tuning_topk}
\end{table*}

\section{Limitations}
This paper highlights the importance and possibility of partial fine-tuning in achieving both model performance and parameter efficiency improvements for the first time. 
While introducing a novel fine-tuned angle metric for guiding the selection of specific layers to be fine-tuned in a given model, it is worth noting that the current approach requires fully fine-tuning the model for several epochs to compute the angle, which incurs additional computational costs prior to partial fine-tuning. Hence, there is ample room for designing a more effective partial fine-tuning strategy. 
Additionally, exploring the validation of partial fine-tuning in the field of natural language processing and investigating the underlying reasons behind its effectiveness could be two promising directions for further exploration.

\end{document}